\pdfoutput=1
\documentclass{article}

\usepackage{microtype}
\usepackage{graphicx}
\usepackage{subfigure}
\usepackage{booktabs} % for professional tables
\usepackage{caption}

\usepackage{hyperref}

\usepackage[accepted]{icml2025}

\usepackage{amsmath}
\usepackage{amssymb}
\usepackage{mathtools}
\usepackage{amsthm}

\usepackage[capitalize,noabbrev]{cleveref}

\theoremstyle{plain}
\newtheorem{theorem}{Theorem}[section]
\newtheorem{proposition}[theorem]{Proposition}
\newtheorem{lemma}[theorem]{Lemma}

\theoremstyle{definition}
\newtheorem{definition}[theorem]{Definition}

\theoremstyle{remark}

\usepackage[textsize=tiny]{todonotes}

\usepackage{verbatim}
\usepackage{amssymb}
\usepackage{bm}
\usepackage{multirow}
\usepackage{amssymb}
\usepackage{graphicx}
\usepackage{float}
\usepackage{overpic}
\usepackage{verbatim}
\usepackage{algorithmic}
\usepackage{algorithm}
\usepackage{colortbl}
\usepackage{wrapfig}
\usepackage{subfigure}  %插入多图时用子图显示的宏包
\usepackage{subcaption}
\newcommand{\etc}{\textit{etc}.}
\newcommand\ie{\textit{i.e.}}
\newcommand\eg{\textit{e.g.}}

\icmltitlerunning{Beyond One-Hot Labels: Semantic Mixing for Model Calibration}

\begin{document}

%\twocolumn[
%\icmltitle{Submission and Formatting Instructions for \\
%           International Conference on Machine Learning (ICML 2025)}
\twocolumn[
% \icmltitle{Exploring Realistic Content Mixup for Balanced Model Calibration}
% \icmltitle{ConfMix: Leveraging Calibration-aware Data Augmentation via Diffusion Models for Model Calibration}
% \icmltitle{Bridging the Calibration Gap with ConfMix: A Data-driven Approach for Model Calibration}
%\icmltitle{Calibrating Models Beyond One-hot Labels: A Calibration-aware Data Augmentation Approach}
\icmltitle{Beyond One-Hot Labels: Semantic Mixing for Model Calibration}
% \icmltitle{ConfMix for Model Calibration: Leveraging Calibration-aware Data Augmentation via Diffusion Models}



\icmlsetsymbol{equal}{*}

\begin{icmlauthorlist}
%Haoyang Luo, Linwei Tao, Minjing Dong, Chang Xu
\icmlauthor{Haoyang Luo}{cityu}
\icmlauthor{Linwei Tao}{usyd}
\icmlauthor{Minjing Dong}{cityu}
\icmlauthor{Chang Xu}{usyd}
\end{icmlauthorlist}

\icmlaffiliation{cityu}{Department of Computer Science, City University of Hong Kong}
\icmlaffiliation{usyd}{School of Computer Science, University of Sydney}

\icmlcorrespondingauthor{Minjing Dong}{minjdong@cityu.edu.hk}

\icmlkeywords{Machine Learning, Model Calibration}

\vskip 0.3in
]



\printAffiliationsAndNotice{}  % leave blank if no need to mention equal contribution

\begin{abstract}
Model calibration seeks to ensure that models produce confidence scores that accurately reflect the true likelihood of their predictions being correct. However, existing calibration approaches are fundamentally tied to datasets of one-hot labels implicitly assuming full certainty in all the annotations. Such datasets are effective for classification but provides insufficient knowledge of uncertainty for model calibration, necessitating the curation of datasets with numerically rich ground-truth confidence values. However, due to the scarcity of uncertain visual examples, such samples are not easily available as real datasets. In this paper, we introduce calibration-aware data augmentation to create synthetic datasets of diverse samples and their ground-truth uncertainty. Specifically, we present \textbf{Calibration-aware Semantic Mixing (CSM)}, a novel framework that generates training samples with mixed class characteristics and annotates them with distinct confidence scores via diffusion models. Based on this framework, we propose calibrated reannotation to tackle the misalignment between the annotated confidence score and the mixing ratio during the diffusion reverse process. Besides, we explore the loss functions that better fit the new data representation paradigm. Experimental results demonstrate that CSM achieves superior calibration compared to the state-of-the-art calibration approaches. Our code is \href{https://github.com/E-Galois/CSM}{available here}.
% Model calibration aims to prevent modern deep models from exhibiting inaccurate confidence levels in their predictions.
%Current methods typically focus on optimization-based calibration techniques, including loss functions, regularization strategies, and post-hoc adjustments. However, these approaches often rely on one-hot, full-confidence labels without accounting for the varying intrinsic hardness of individual samples. 
% In this paper, we present \textbf{ConfMix}, \textcolor{blue}{a novel framework that enhances confidence calibration by creating and annotating sample variants with distinct levels of difficulty. To this end, we aim to define an inter-class transform bridge that generates hard samples associated to intermediate semantics. By utilizing diffusion models, single-label source images are semantically transformed with a range of guidance strengths to populate a continuous spectrum of hardness levels. Since the optimal confidence for a sample can vary across models and human evaluators, we cannot adopt fixed hardness levels directly as confidence estimation. From the perspective of sample hardness, we focus on aligning the model-estimated hardness with a debiased hardness annotation for accurate calibration. To facilitate this alignment, we propose a class-aware measurement for precise hardness differentiation and leverage CLIP-based annotations to evaluate the hardness of augmented images.} Experimental results demonstrate that ConfMix achieves superior calibration compared to state-of-the-art calibration approaches.
\end{abstract}

\section{Introduction}
\label{sec:intro}

Modern deep neural networks (DNNs) have achieved significant effectiveness in various vision tasks including image recognition, retrieval, and object segmentation \cite{zagoruyko2016wide,gordo2016deep,kirillov2023segment}. 
Although DNNs can achieve exceptional accuracy, they could be unreliable in real-world scenarios since the predictions are always over-confident \cite{FLSD,wei2022mitigating}. This issue could mislead security-sensitive applications, such as autonomous systems, surveillance applications, medical diagnostics, \etc~Model calibration \cite{vaicenavicius2019evaluating, tao2023benchmark}, the process which refines the models' predicted confidence to be consistent with the actual probability of correct prediction, stands as the key to reliable vision models.

% 1. Attribute the issue of overconfidence to optimization on hard labels. However, existing algorithms focus on an ordinary data structure. We are motivated to explore data-driven calibration methods.
Various calibration techniques have been introduced, including post-hoc calibration \cite{platt1999probabilistic,ECE,tao2025feature}, loss function designs \cite{FLSD,dual_focal}, and regularization methods \cite{kumar2018trainable,krishnan2020improving}. 
However, current methods rely on one-hot labeled data, which inaccurately assume uniform uncertainty across samples for confidence estimation. For instance, while a clear cat image and an ambiguous cat-dog hybrid image demand distinct confidence distributions, one-hot labels enforce identical certainty. This creates a critical gap that current datasets lack ground-truth uncertainty annotations necessary to teach models nuanced confidence distinctions. However, real-world collection of such data is infeasible due to scalability and ambiguity challenges. This necessitates us to create synthetically generated datasets with diverse ground-truth confidence annotations.

\begin{figure}[t]
\vskip 0.2in
\begin{center}
\centerline{
\includegraphics[width=0.99\columnwidth]{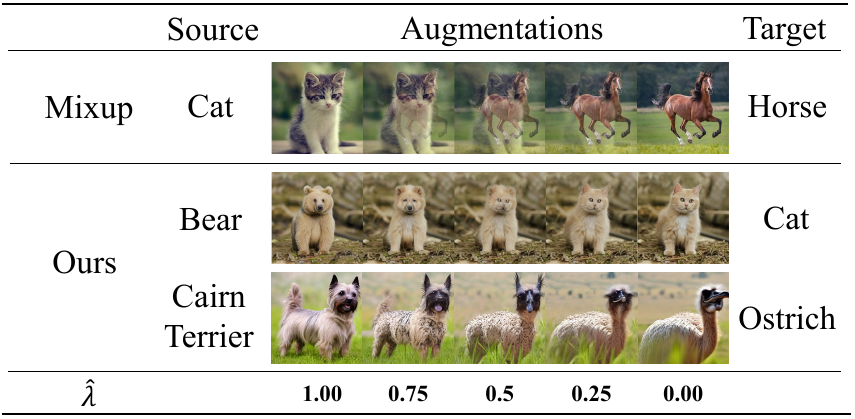}
}
\caption{\textbf{Differences between Mixup and the Proposed Augmentation.} Samples with intermediate mixing coefficients exhibit simple overlapping pattern for Mixup. The proposed augmentation mixes semantics while preserving the object completeness, displaying enhanced image fidelity.}%\vspace{-3ex}
\label{fig:motivation}
\end{center}
\vskip -0.2in
\end{figure}

% 2. Why not mixup? (discuss mixup related work and its limitations)
%To generate training samples for calibration, a naive solution could be Mixup-based approaches \cite{yun2019cutmix, kim2020puzzle, uddin2020saliencymix, chen2022transmix, noh2023rankmixup}. 
To generate samples with ground-truth uncertainty, a naive solution could be Mixup-based approaches \cite{yun2019cutmix, kim2020puzzle, uddin2021saliencymix, chen2022transmix, noh2023rankmixup}. 
By augmenting training data with pixel-interpolated samples and soft labels, DNNs are supposed to learn diverse confidence. However, such methods often produce low-fidelity samples due to unrealistic pixel fusion or fragmentary collage, which deviate significantly from the distribution of real-world data as shown in \cref{fig:motivation}. 
Consequently, these approaches struggle to generalize the learned calibration knowledge to real-world scenarios and hardly contribute to the calibration improvement \cite{wen2020combining,maronas2021calibration,wang2023pitfall}. 
Data augmentations have also been explored at test time \cite{hekler2023test} to improve uncertainty estimation. While these post-hoc methods can mitigate over-confidence globally, they often fail to achieve precise calibration because they ignore miscalibration in the training with one-hot samples.

% 3. Introduce your algorithm
For successful calibration-aware data augmentation, realistic samples with different soft class posteriors are expected to be constructed, which remains unexplored in the field of calibration. In this paper, we introduce Calibration-aware Semantic Mixing (CSM), a data augmentation approach designed for model calibration that generates high-fidelity samples with semantic mixing. 
% introduce 3.1 
Specifically, we adopt pretrained diffusion models to sample different sets of augmented images. %Within the same set, the generated images are conditioned on the same latent noise but different soft class posteriors, which ensures intra-set layout consistency and high-fidelity semantic mixing. 
Within the same set, the generated images are conditioned on the same latent noise but different soft class posteriors. Therefore, objects can be continuously transformed in the pixel layout from the instance of one class to another, which preserves the basic spatial completeness of objects to ensure image fidelity.

% introduce reannotation
Besides these image samples, their corresponding soft labels should be annotated for learning. A naive solution is the mixing ratio during the diffusion reverse process. However, it can rarely indicate the class posterior precisely.
%To tackle this issue, we propose to reannotate the data by leveraging CLIP's visual features, which exhibit reduced bias due to normalized pretraining. 
To tackle this issue, we propose to reannotate the data by leveraging CLIP's visual features which have been jointly trained with massive language supervision \cite{CLIP}, and are thus able to provide more precise class posteriors. 
%By representing the mixing coefficient through optimized sample embeddings, which are further expressed as interpolations of class prototypes, 
By representing the optimized sample embeddings as interpolations of class prototypes, we can identify and eliminate class-specific biases in the estimated mixing coefficients, resulting in more accurate and effective annotations.
% introduce loss function
Moreover, we highlight the problem of imbalanced fitting of augmented data when utilizing traditional loss functions for calibration through some theoretical analysis. We reveal that the training of augmented data with $\mathcal{L}_2$ loss intrinsically results in balanced learning and better calibration performance. 
% experiments
Extensive experiments on various benchmarks and tasks demonstrate our method's strong effectiveness in accurately estimating confidence levels, achieving superior results for data-aware model calibration.

\section{Related Work}
\label{sec:rel_work}

\subsection{Model Calibration}
Network calibration \cite{ECE} requires techniques to align model confidence with actual accuracy. Post-hoc methods adjust test-time parameters using a validation set to improve calibration. Notable techniques include Temperature Scaling (TS) \cite{ECE}, Histogram Binning \cite{zadrozny2001obtaining}, Beta calibration \cite{kull2017beta} and its extension to Dirichlet calibration \cite{kull2019beyond}. While simple, post-hoc methods are sensitive to distributional shifts \cite{ovadia2019can}. Training-time regularization methods involve explicit learning constraints and implicit loss functions. Label Smoothing (LS) \cite{muller2019does} reduces output entropy by replacing one-hot targets with smoothened labels. Margin-based Label Smoothing (MbLS) \cite{liu2022devil} balances model discriminability and calibration by imposing a margin on logits. Implicit loss functions, such as those optimizing Expected Calibration Error \cite{karandikar2021soft, kumar2018trainable}, Focal loss \cite{focal_loss, FLSD}, Inverse Focal loss \cite{wang2021rethinking}, and Mean Square Error \cite{hui2020evaluation, liang2024calibrating}, are effective objectives to improve calibration. %can also improve calibration but may lead to overconfidence on datasets with one-hot labels. 
Data-aware calibration primarily involves perturbation methods \cite{tao2024consistency} and Mixup-based methods \cite{zhang2018mixup, yun2019cutmix, hendrycks2019augmix}, originally designed for improved generalization . These techniques have been found effective for calibration \cite{zhang2018mixup, pinto2022using}. RegMixup \cite{pinto2022using} uses Mixup as a regularizer for cross-entropy loss, enhancing both accuracy and uncertainty estimation. RankMixup \cite{noh2023rankmixup} incorporates ordinal ranking relationships among samples to reduce Mixup label bias. However, recent work \cite{wang2023pitfall} highlights the limitation of simple convex combinations in Mixup methods.

\subsection{Generative Data Augmentation}
\begin{comment}
Recently, generative methods such as Diffusion Models (DM) are utilized for data augmentation thanks to its advanced fidelity. To improve image diversity, label-preserving augmentations based on DMs are proposed. Real-Guidance \cite{he2022synthetic} performs augmentation on scarce data by generation in-class images with diverse class descriptions.  DA-Fusion \cite{trabucco2023effective} leverages SDEdit \cite{meng2021sdedit} technique to generate image variations of controlled difference to produce within-category augmentations. Diff-Mix \cite{wang2024enhance} perform diffusion-enhanced image
translations by transferring the foreground object to another class while preserving the image background. In contrast, DiffuseMix \cite{islam2024diffusemix} alters the background with crafted prompts, such as ``Autumn," to enhance the style diversity. To truly mix the class concepts, \cite{MagicMix} combins SDEdit with specified prompts to generate semantically mixed images. \cite{chen2024decoupled} propose to decouple class-dependent parts with class-independent parts in images and reblend them to generate one-hot or mixed-label augmentations. Although semantic mixing has been explored for enhancing classification accuracy, limited works have investigated its effectiveness for calibration, motivating us to propose our CSM.
\end{comment}
Generative methods, particularly Diffusion Models (DM), have recently gained popularity for data augmentation due to their high fidelity \cite{he2022synthetic}. Label-preserving augmentations based on DMs have been proposed to enhance image diversity. DA-Fusion \cite{trabucco2023effective} uses the SDEdit \cite{meng2021sdedit} technique to produce controlled variations of images within the same class. Diff-Mix \cite{wang2024enhance} performs image translations by transferring foreground objects to a different class while preserving the background. In contrast, DiffuseMix \cite{islam2024diffusemix} modifies the background with crafted prompts to enhance style diversity. To mix class concepts, MagicMix \cite{MagicMix} combines SDEdit with specific prompts to generate semantically mixed images. De-DA \cite{chen2024decoupled} decouples and re-blends class's dependent and independent parts to create one-hot or mixed-label augmentations. %While semantic mixing has been explored to improve classification accuracy, few studies have examined its impact on calibration, motivating the development of our CSM.

% 3.1 preliminaries (Calibration (ece), diffusion models)
% 3.2 Introduce ConfMix. 
% a. To begin with, introduce motivation: existing calibration techniques focus on loss/regularization, etc., and the influence of training data on calibration remains unexplored. -> Discuss the limitations of existing datasets (one-hot), and discuss why mixup does not work here -> our objective: high-fidelity images with soft labels.
% b. How to achieve our objective? -> Our solution (ConfMix), introduce how to generate images with soft labels in detail with equations and defined variables.
% 3.3 Introduce XXXXX
% a. ConfMix can provide high-fidelity images with soft labels. However, these soft labels might not be accurate. -> We need reannotation.
% b. The optimization needs improvement since the labels are soft instead of one-hot format. Why previous loss functions do not work here? How to improve?
\section{Method}
\label{sec:method}
\begin{figure*}[t]
\centering
\includegraphics[width=\linewidth]{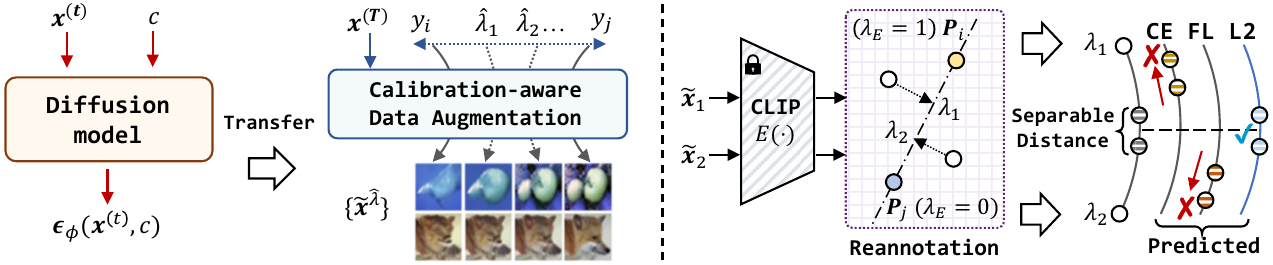}
\caption{\textbf{The framework of the proposed Calibration-aware Semantic Mixing (CSM).} Left: Calibration-aware Data Augmentation exploits a pretrained diffusion model to ensure fidelity and coherence in generating conceptually mixed augmentations. Right: To further improve the annotation and fitting of class posteriors, a normalized reannotation scheme and the theoretically unbiased L2 loss are adopted to achieve balanced confidence calibration.}
\label{fig:framework}
\end{figure*}

We provide preliminaries of model calibration and diffusion models in \cref{sec:preliminaries} and then introduce our proposed framework illustrated in \cref{fig:framework}, regarding details of semantic mixing, reannotation, and learning objective for model calibration in \cref{sec:EMS} and \cref{sec:CRL}.

% and then introduce ConfMix framework in \cref{sec:EMS} and XXX in \cref{sec:CRL}.

\subsection{Preliminaries}
\label{sec:preliminaries}
\textbf{Model Calibration}\quad Considering an image classification task with a dataset $\{(\bm x_i, y_i)\}_{i=1}^{N}$, where $\bm x_i$ is a sample and $y_i$ is its true label, a DNN classifier $f_\theta$ predicts class probabilities $\bm p_i = \{p_i^k\}_{k=1}^K = \operatorname{softmax}(f_\theta(\bm x_i)) \in \Delta^K$, with $p_i^k$ representing the predicted probability for class $k$. Here, $\Delta^K=\big\{\bm p \in [0,1]^K,\sum_{k=1}^{K}p_k=1\big\}$ represents the $K$-dimensional probability space. The predicted label for sample $i$ is% the highest predicted probability:
\begin{align}
\hat{y}_i = \mathop{\arg\max}\limits_{k}~p_i^k.
\label{eq:prediction}
\end{align}
The confidence score of a predicted label $\hat{y_i}$ is defined as $\hat{p}_i=p_i^{\hat{y_i}}$.
Formally, a model is well-calibrated if the predicted confidence matches the true accuracy, \ie, $\mathbb P(\hat{y}_i = y_i \mid p_i^{\hat{y_i}}=p)$ for $p \in [0,1]$.
To measure network calibration, the Expected Calibration Error (ECE) is a commonly-used metric defined as $\mathbb E_{\hat{p}_i}[\mathbb P(\hat{y_i}=y_i\mid \hat{p}_i)-\hat{p}_i]$. Practically, due to the finite samples available, ECE is approximated as the weighted average of the absolute difference between accuracy and confidence across $M$ bins of grouped confidence values. Formally, ECE is defined as:
\begin{align}
\text{ECE} = \frac{1}{M} \sum_{m=1}^{M} |B_m| \cdot | \text{Acc}(B_m) - \text{Conf}(B_m) |,
\label{eq:ECE}
\end{align}
where $\text{Acc}(B_m)$ and $\text{Conf}(B_m)$ denote the accuracy and average confidence in the $m$-th bin, and $|B_m|$ is the number of samples in $B_m$.

% \noindent\textbf{Mixup}\quad An effective model calibration scheme is to introduce auxiliary samples through mixup \cite{zhang2018mixup} augmentations, which generates fused samples and labels from random training pairs as
% \begin{align}
% \left \{
% \begin{array}{l}
% \widetilde{\bm x} = \lambda \bm x_{i} + (1 - \lambda) \bm x_{j}\\
% \widetilde{\bm q} = \lambda \bm q_{i} + (1 - \lambda) \bm q_{j}
% \end{array}
% \right.
% ,
% \label{eq:mixup}
% \end{align}
% where $\lambda\sim \operatorname{Beta}(\alpha, \alpha) \in [0,1]$ is a mixup coefficient. Generally, mixup methods manifest the model confidence by fitting such soft label distributions $\widetilde{q}$. However, they inevitably produce low-fidelity off-domain samples and yield semantically-ambiguous fusions due to the limited knowledge in pixel space. 
% %decreasing the reliability of $\lambda$-parameterized labels.

\noindent\textbf{Diffusion Models for Image Generation}\quad Diffusion models are trained to generate realistic images via a gradual denoising process upon Gaussian noises. Their forward process gradually adds noise through a Markov chain with Gaussian transitions $q_{\text{fwd}}(\bm x_{i}^{(t)} | \bm x_{i}^{(t-1)}) = \mathcal{N}(\bm x_{i}^{(t)}; \sqrt{\eta_t}\bm x_{i}^{(t)}, (1-\eta_t)\bm I)$, where $\bm x_{i}^{(t)}$ is the noised variable of $\bm x_{i}^{(0)} = \bm x_{i}$ at step $t$ and $\eta_t$ is the noise schedule. The objective for training conditional denoising parameters $\phi$ is represented as $\min_{\phi} \mathbb{E}_{\bm \epsilon, \bm x, y, t} \Vert \bm\epsilon-\bm\epsilon_{\phi}^{t}(\bm x^{(t)}, y) \Vert$, where $\bm\epsilon_{\phi}$ is the predicted noise. For conditional image generation, the classifier-free guidance (CFG) \cite{CFG} is adopted to denoise images altering noise prediction as
\begin{align}
\widetilde{\bm\epsilon}_{\phi}(\bm x^{(t)}, y) = (1+\omega)\bm\epsilon_{\phi}(\bm x^{(t)}, y) - \omega\bm\epsilon_{\phi}(\bm x^{(t)}, \varnothing),
\label{eq:CFG}
\end{align}
where $\omega$ denotes the guidance strength.

\subsection{Calibration-aware Data Augmentation}
\label{sec:EMS}
% 1. Existing algorithms tackle the calibration issue from different perspectives, XXXX.
% 2. However, model training is performed on datasets with an ordinary structure. -> model tends to be overconfident since the labels are in one-hot format.
% 3. In this paper, we propose to tackle calibration from a data perspective. -> mixup could be a naive solution. (introduce mixup and discuss limitations)
% 4. Our objective: XXX
% 5. Introduce your framework
%Various calibration techniques have been introduced to improve the reliability of model predictions from different perspectives, such as post-hoc calibration \cite{ECE}, loss functions \cite{FLSD,dual_focal}, ensemble \cite{zhang2020mix}, \etc~However, 
Existing calibration approaches \cite{ECE, FLSD, kumar2018trainable, muller2019does} primarily focus on post-hoc and training techniques, while the training data for calibration is rarely considered.
% incorporate various regularization techniques into the model optimization process,
In fact, the training is always performed on the datasets with an ordinary structure, \ie, the data pair of an image and its corresponding single label in the image classification task. Such annotated data exhibit zero uncertainty which can be ineffective for confidence calibration. %All the data pairs are provided in an ``overconfident" manner since the image class is annotated as a hard label, \ie, one-hot encoding format with full confidence, which is far away from the accuracy in reality. To mitigate this gap, we propose to tackle model calibration from a unique data perspective via calibration-aware data augmentation. 
To obtain diverse samples of ground-truth uncertainty values, we propose to tackle model calibration via a calibration-aware data augmentation scheme. 
%先定义出来我们不同于one-hot label，我们这里想生成soft label让模型学习对应的confidence。
Specifically, soft-labeled data are defined as $\{(\widetilde{\bm x}_a, \widetilde{\bm q}_a)\}_{a=1}^{A}$, where $A$ is the total number of augmented samples and $\{\widetilde{q}_a^{k}\}_{k=1}^K \in \Delta^K$ can be the non-binary confidence value for each class $k$ of sample $\widetilde{\bm x}_a$. We expect such data to provide confidence knowledge via soft label $\widetilde{\bm q}_a$ instead of single class $y$ in order to calibrate the model. 
%先简单讲下mixup是个naive solution，讲mixup的时候，结合3.1里面已有的定义好的公式和变量，把mixup的公式3-4句话放段落里一起讲一下。

A straightforward calibration scheme generating soft-labeled data is the mixup \cite{zhang2018mixup} augmentation, which generates fused samples and labels from random training pairs as $\widetilde{\bm x} = \lambda \bm x_{i} + (1 - \lambda) \bm x_{j}$ and $\widetilde{\bm q} = \lambda \bm q_{i} + (1 - \lambda) \bm q_{j}$, where $\lambda\sim \operatorname{Beta}(\alpha, \alpha) \in [0,1]$ is a mixup coefficient, $\alpha > 0$ is a hyperparameter controlling the interpolation strength \cite{noh2023rankmixup}, and $\bm q_{i}$ is the one-hot vector of $y_i$.
%然后讲为什么它在这里不work。
%Mixup imposes smooth transitions around the decision boundaries, leading to measurable confidence levels for model calibration. 
However, the direct utilization of mixup cannot contribute to the calibration improvement, as empirically found by several studies \cite{wen2020combining,maronas2021calibration,wang2023pitfall}. We mainly attribute this failure to the fact that mixup performs calibration learning on low-fidelity images due to simple convex combinations, which makes the learned calibration difficult to generalize to real-world data.
% However, mixup is practically a naive solution for calibration. It mainly produces low-fidelity images, which can inevitably drive the real learning process off-domain.
% Meanwhile, it is challenging to blend object instances according to an assumed ratio due to not only the semantical integrity of entire objects in the pixel space, but also the complexity in measuring such ratio \cite{chen2022transmix}.
%As a result, common mixup methods often suffer from generating off-domain sample series \cite{yun2019cutmix, kim2020puzzle, uddin2020saliencymix} that are unrealistic pixel collages, degrading their guidance for calibration \cite{noh2023rankmixup}. 
%Figure \ref{fig:bad_mixup} illustrates the issues when learning with these low-quality augmentations and raises a natural question: How can we generate highly vicinal samples that transfer in-domain from one class to another?
%再引出你的目标是什么。
Therefore, to establish an effective calibration-aware data augmentation, we propose to generate high-fidelity images with semantic mixing, which reflects the shifting process of class posteriors from one category to another.
% Therefore, to establish an effective calibration-aware data augmentation, two key questions arise: 1) \textbf{Fidelity}: How can we generate conceptually authentic sample series that respect varying mixing ratios? 2) \textbf{Coherence}: How can these series seamlessly transfer within-domain, ensuring the integrity of objects as they shift from one category to another?

We start by expressing the general problem as how to establish a vicinal distribution that can be sampled from, \ie,
\begin{align}
&\widetilde{\bm x} \sim p_v(\widetilde{\bm x} \mid \bm x_i, \bm x_j)\nonumber,\\
&\text{s.t.}~(\bm x_i,\bm x_j) \sim p_{dual}(\bm x_i,\bm x_j \mid y_i, y_j),
\label{eq:general_problem}
\end{align}
where $p_{dual}$ is the joint distribution for $\bm x_i$ and $\bm x_j$ and $p_v$ is the vicinal distribution to be designed for semantic mixing. 
%Given data distribution $p(\bm x~|~c)$, pixel-level mixups typically implement this idea as $\widetilde{\bm x}=\lambda F_{pix}(\bm x_i) + (1-\lambda)F_{pix}(\bm x_j)$ and $\bm x_i \sim p(\bm x\mid c_i)$, where $F_{pix}$ is a designed mapping or masking function and $\lambda \in [0, 1]\sim \operatorname{Beta}(\alpha, \alpha)$. Though simple, these methods are intrinsically vulnerable to yield unrealistic off-domain images due to the limited semantical manageability within pixels. Meanwhile, the joint distribution of $\bm x_i$ and $\bm x_j$ is basically considered as the uncorrelated product distribution, undermining its potential for coherent object fusion with intermediate $\hat{\lambda}$s.

%Considering these factors, 
To blend class semantics with soft labels, we propose a dual-conditioned sampling strategy via a conditional diffusion model explicitly optimized to estimate $p(\bm x \mid y)$, \textit{i.e.}, 
\begin{align}
p_{\phi}(\bm x \mid \bm q^y)=\int_{\bm x^{(T)}}p_{\phi}(\bm x \mid %\bm q_y
\bm q^y,\bm x^{(T)})p(\bm x^{(T)})\mathrm{d} \bm x^{(T)},
\label{eq:conditional_dm}
\end{align}
Specifically, we generate augmented samples $\{\widetilde{\bm x}^{\hat{\lambda}}\}$ via diffusion reverse process as
\begin{align}
&\widetilde{\bm x}^{0} \sim p_{\phi}(\bm x \mid \bm q^{y_j},\bm x^{(T)}),\quad
\widetilde{\bm x}^{1} \sim p_{\phi}(\bm x \mid \bm q^{y_i}, \bm x^{(T)}),\nonumber\\
&\widetilde{\bm x}^{\hat{\lambda}} \sim p_{\phi}(\bm x \mid \widetilde{\bm q}^{\hat{\lambda}},\bm x^{(T)}),\quad
\bm x^{(T)} \sim \mathcal{N}(\bm x^{(T)}, 0, \bm I),
\label{eq:our_solution}
\end{align}
where $\hat{\lambda} \in [0,1]$. The augmentation is simultaneously conditioned on the mixed label $\widetilde{\bm q}^{\hat{\lambda}}=\hat\lambda \bm q^{y_i} + (1-\hat\lambda)\bm q^{y_j}$ and the sampled Gaussian noise $\bm x^{(T)}$ at diffusion timestep $T$.
The merits here are 3-fold: 1) the data fusion introduced at diffusion conditioning rather than pixel levels can significantly improve the image fidelity, which aligns with the real data distributions. 2) $\widetilde{\bm x}^{0}$, $\widetilde{\bm x}^{1}$, and $\widetilde{\bm x}^{\hat{\lambda}}$ are correlated by conditioning on a common hidden state $\bm x^{(T)}$, which controls the general image layout, maximizes the proximity of the augmentation series, and ensures unified visual composition for calibration learning. 3) conditional diffusion models are implicit class posterior estimators \cite{li2023your} that can generalize to unseen conditions \cite{li2024understanding}, \textit{i.e.}, mixed labels in our case.
% , enhancing the results' alignment to the mixing ratio $\hat{\lambda}$.
Therefore, $\widetilde{\bm q}^{\hat{\lambda}}$ can be a naive solution to the label of augmented sample $\widetilde{\bm x}^{\hat{\lambda}}$.

\subsection{Calibrated Reannotation and Balanced Learning}
\label{sec:CRL}
% 1. Based on the framework introduced in 3.2, discuss the potential issues.
% a. The $\lambda$ can be the annotated labels, but they could be inaccurate. Why? Provide theoretical evidence or empirical observations.
% b. Now we have data pair of generated images with semantic fusion and their soft labels, but existing popular loss functions cannot optimize this type of data pair correctly. Why? Provide theoretical evidence or empirical observations.
% 2. Introduce calibrated reannotation and balanced learning.
While the proposed data augmentation scheme in \cref{sec:EMS} demonstrates various advantages for calibration learning, several potential issues arise when considering the characteristics of semantic mixing data pair $(\tilde{x},\tilde{q})$.
% reannotation
One issue lies in the annotation. Given a semantic-mixed image sample $\widetilde{x}^{\hat{\lambda}}$, it is necessary to provide an accurate annotation of its true class posterior. However, the annotated soft label $\widetilde{\bm q}^{\hat{\lambda}}$ from diffusion models can hardly be equivalent to the true class posterior since diffusion models are also pretrained on the datasets with hard labels. In fact, we empirically found that the generated sample set $\{\widetilde{x}^{\hat{\lambda}}\}$ where $\hat{\lambda} \in [0,1]$ shows high proximity, which strengthens the layout and object coherence due to the joint conditioning on $\bm x^{(T)}$. As shown in \cref{fig:visualization_sample_lambda}, $\hat{\lambda}$ may not accurately reflect the true class posterior due to non-linear semantic transitions in generated images.
% balanced learning
%Another issue lies in the training phase since existing loss functions are designed for samples with hard labels. 
Another issue arises in the training phase, as the existing loss functions are specifically designed for hard labels. 
We provide some theoretical analysis of the potential bias induced by popular loss functions used in the calibration field during the training of soft-labeled data. 
% our solution
To tackle these issues, we introduce calibrated reannotation and balanced learning to ensure effective calibration learning.

\textbf{Calibrated Reannotation}\quad
%上一段的分析：One issue lies in the annotation. Given a semantic-mixed image sample $\widetilde{x}^{\hat{\lambda}}$, it is necessary to provide an accurate annotation of its true class posterior. However, the annotated soft label $\widetilde{\bm q}_{\hat{\lambda}}$ from diffusion models can hardly be equivalent to the true class posterior since diffusion models are also pretrained on the datasets with hard labels. In fact, we empirically found that the generated sample set $\{\widetilde{x}^{\hat{\lambda}}\}$ where $\hat{\lambda} \in [0,1]$ shows high-proximity, which strengthens the layout and object coherence due to the joint conditioning on $\bm x^{(T)}$. As shown in XXXX, .
% 1. 接着上一段的分析，3.2里面的\hat{lambda}不够好。那么我们现在要重新reannotate这个soft label。先提出我们的假设：有一个在augmented sample with semantic mixing上达到最优的visual encoder。
% 2. 然后是你的公式6，算出最优的lambda。再relax 条件，引入lambda_E。
% 3. 公式8到9的过度感觉逻辑不对。你写的是因为公式8里面有bias项，所以我们去掉这一项，但是公式8是你把公式7往公式6里面代算出来的，而公式6算的是最优的lambda值，那公式8这里也应该是最优值？你相当于对最优值做了估计，那为什么不直接用公式6？
% 4. 我感觉你这里的逻辑可能可以这样：公式6确实是最优值，但是对E的假设太强了，所以你relax对E的假设，引入了公式7，引入了lambda_E去做估计。但这里你得好好说明下lambda_E的作用。代入后，推出公式8，发现lambda的公式中多了bias，我们提出将bias项去掉，以得到更加精确的估计。
As analyzed, $\hat{\lambda}$ is inaccurate as class posterior estimation, which motivates us to reannotate these augmented samples with enhanced soft labels.
Aiming for precise $\lambda$ annotation, we assume that there exists a classification-optimal visual encoder $\operatorname{E}(\cdot)$ on the dataset and our semantic-mixed augmented samples. In this case, the \textit{confidence ratio} of the two mixing classes $i$ and $j$ for the same sample $\widetilde{\bm x}$ should be
%Due to the variance in data itself, models that are either generatively or discriminatively learned can all produce embeddings of distinctive norms, which can yield poor model calibration \cite{wei2022mitigating}. 
%However, such behavior is rarely seen with CLIP since it's innately learned with normalized features. Although it's now hard to determine the true mixing coefficient $\lambda$ for the augmentations in the pixel space, the fusion strength can instead be measured within CLIP's semantic space. Assuming that CLIP's visual encoder $\operatorname{E}(\cdot)$ is generally classification-optimal on our augmented samples, the \textit{confidence ratio} of the two mixing classes $c_i$ and $c_j$ for the same sample $\widetilde{\bm x}$ should be
\begin{align}
&\exp(\frac{\operatorname{E}(\widetilde{\bm x})^\top\bm P_{i} - \operatorname{E}(\widetilde{\bm x})^\top\bm P_{j}}{\tau})=\frac{\lambda}{1-\lambda},\nonumber\\
&\Leftrightarrow \lambda=\sigma(\frac{\operatorname{E}(\widetilde{\bm x})^\top\bm P_{i} - \operatorname{E}(\widetilde{\bm x})^\top\bm P_{j}}{\tau}),
\label{eq:probability_ratio}
\end{align}
where $\bm P_{k}=\sum_{y_i=k} \operatorname{E}(\bm x_i)$ is the class prototype for class $k$, $\sigma$ is the sigmoid function, $\lambda$ is the reannotated mixing coefficient replacing $\hat{\lambda}$, and $\tau$ is a temperature parameter. 
%公式6确实是最优值，但是对E的假设太强了，所以你relax对E的假设，引入了公式7，引入了lambda_E去做估计。
Eq. \eqref{eq:probability_ratio} is the theoretically optimal $\lambda$ annotation. However, the assumption of optimal $\operatorname{E}(\cdot)$ can be overly strong in practice. Therefore, we first introduce a feature-level $\lambda$-annotation, \ie $\lambda_E$, to relax the encoded feature as
%It's nonequivalent to regard arbitrary $\operatorname{E}(\widetilde{\bm x})$ as the mixup of $\bm P_{i}$ and $\bm P_{j}$ as in vanilla feature mixup. Nevertheless, we can relax the encoded feature and acquire a feature-level mixing coefficient $\lambda_E$ through 
\begin{align}
&\operatorname{E}(\widetilde{\bm x})=\lambda_E\bm P_{i}+(1-\lambda_E)\bm P_{j} + \bm r,\nonumber\\
&s.t.~\bm r^\top(\bm P_{i}-\bm P_{j}) = 0,
\label{eq:feature_expressed}
\end{align}
where $\lambda_E$ serves as the interpolation coefficient and $\bm r$ is a linearly independent residual term.
%In this way, $\lambda_E$ is only measured by the affine space defined by the class prototypes regardless of other biased factors as we'll see later after \cref{eq:lambda_expressed}. 
Here, $\lambda_E$ is acquired by projecting $\operatorname{E}(\widetilde{\bm x})$ onto the 1-D affine space of $\bm P_i$ and $\bm P_j$. Specifically, by taking the inner product of both sides of \cref{eq:feature_expressed} with $\bm P_i -\bm P_j$, we can express $\lambda_E$ as
%但这里你得好好说明下lambda_E的作用。最好再公式7后面解释下这个新引入的变量以及意义。以及lambda_E是怎么算出来的。
\begin{align}
&\lambda_E=\frac{(\operatorname{E}(\widetilde{\bm x})-\bm P_{j})^\top(\bm P_{i}-\bm P_{j})}{(\bm P_{i}-\bm P_{j})^2},
\label{eq:lambda_E_expressed}
\end{align}
\begin{comment}
Therefore, $\lambda_E$ can be expressed as 
\begin{align}
&\lambda_E=\frac{(\operatorname{E}(\widetilde{\bm x})-\bm P_{j})^\top(\bm P_{i}-\bm P_{j})}{(\bm P_{i}-\bm P_{j})^2},
\label{eq:lambda_E_expressed}
\end{align}
which is independent of $\bm r$. 
\end{comment}
Meanwhile, Eq. \eqref{eq:probability_ratio} can be reformulated by substituting Eq. \eqref{eq:feature_expressed} into Eq. \eqref{eq:probability_ratio} as
\begin{comment}
\begin{align}
\lambda=\sigma\Big(\frac{1}{\tau}\Big((\bm P_{i}-\bm P_{j})^2\lambda_E+\bm P_{i}^\top\bm P_{j}-\bm P_{j}^2)\Big)\Big).
\label{eq:lambda_expressed}
\end{align}
\end{comment}
\begin{align}
\lambda=\sigma\Big(\frac{1}{\tau}\Big((\bm P_{i}-\bm P_{j})^2(\lambda_E-\frac{1}{2})+\frac{1}{2}(\bm P_{i}^2-\bm P_{j}^2)\Big)\Big).
\label{eq:lambda_expressed}
\end{align}
See \cref{sec:proofs_reanno} for proof of the above Equations. 
It is evident that directly adopting Eq. \eqref{eq:lambda_expressed} to calculate $\lambda$ would induce bias due to the unequal class norms in $\bm P_{i}^2-\bm P_{j}^2 \ne 0$ and inter-class distances in $(\bm P_{i}-\bm P_{j})^2 \ne (\bm P_{k}-\bm P_{l})^2$ for $i \ne j$ and $\{i,j\}\ne\{k,l\}$, respectively. Note that $\lambda_E$ is only measured by the affine space of the class prototypes regardless of these biases. We propose to regard these factors as invariant across classes and simplify Eq. \eqref{eq:lambda_expressed} as
\begin{align}
\lambda=\sigma\big(s\cdot(\lambda_E-\frac{1}{2})\big),
\label{eq:lambda_rewrite}
\end{align}
where $s > 0$ is a scaling factor. With Eq. \eqref{eq:lambda_rewrite}, the new annotations are free from biases inherent in the visual encoder $\operatorname{E}(\cdot)$. 
In practice, we adopt CLIP's visual encoder as $\operatorname{E}(\cdot)$. 

\begin{table*}[t]
%\begin{center}
\caption{\textbf{Calibration errors before and after temperature scaling.} Results in the brackets are post-temperature results. R50: ResNet-50, R101: ResNet-101. Available results on the same used settings are cited from \cite{noh2023rankmixup}.} \label{table_main}
\vskip 0.1in
\begin{center}
\begin{sc}
\resizebox{1.0\textwidth}{!}{
\begin{tabular}{lcccccc|cccccc|cccccc}
\toprule
\multicolumn{1}{c}{\multirow{3}{*}{Method}} & \multicolumn{6}{c}{CIFAR10 \cite{cifar}}                                                                                  & \multicolumn{6}{c}{CIFAR100 \cite{cifar}}                                                                                & \multicolumn{6}{c}{Tiny-ImageNet \cite{tiny}}                                                                           \\
\multicolumn{1}{c}{}                        & \multicolumn{3}{c}{R50 \cite{resnet}}                & \multicolumn{3}{c}{R101 \cite{resnet}}               & \multicolumn{3}{c}{R50 \cite{resnet}}                & \multicolumn{3}{c}{R101 \cite{resnet}}               & \multicolumn{3}{c}{R50 \cite{resnet}}                & \multicolumn{3}{c}{R101 \cite{resnet}}               \\
\multicolumn{1}{c}{}                        & ACC$\uparrow$    & ECE$\downarrow$       & AECE$\downarrow$      & ACC$\uparrow$    & ECE$\downarrow$       & AECE$\downarrow$      & ACC$\uparrow$    & ECE$\downarrow$       & AECE$\downarrow$      & ACC$\uparrow$    & ECE$\downarrow$       & AECE$\downarrow$      & ACC$\uparrow$    & ECE$\downarrow$       & AECE$\downarrow$      & ACC$\uparrow$    & ECE$\downarrow$       & AECE$\downarrow$      \\ 
\midrule
CE                                          & 95.38          & 3.75(0.97)          & 2.98(1.01)          & 94.46          & 3.61(0.92)          & 3.55(0.85)          & 77.81          & 13.59(2.93)         & 13.54(2.86)         & 77.48          & 12.94(2.63)         & 12.94(2.66)         & 64.34          & 3.18(3.18)          & 2.87(2.87)          & 66.04          & 3.50(3.50)          & 3.52(3.52)          \\
MMCE                                        & 95.18          & 3.88(0.97)          & 3.88(1.12)          & 94.99          & 3.88(1.15)          & 3.88(12.9)          & 77.56          & 12.72(2.83)         & 12.71(2.86)         & 77.82          & 13.43(3.06)         & 13.42(2.80)         & 64.80          & 2.03(2.03)          & 1.97(1.97)          & 66.44          & 3.40(3.40)          & 3.38(3.38)          \\
ECP                                         & 94.75          & 4.01(1.06)          & 3.99(1.53)          & 93.97          & 4.41(1.72)          & 4.40(1.70)          & 76.20          & 12.29(2.08)         & 12.28(2.22)         & 76.81          & 13.43(2.92)         & 13.42(3.04)         & 64.88          & 1.94(1.94)          & 1.95(1.95)          & 66.20          & 2.72(2.72)          & 2.70(2.70)          \\
LS                                          & 94.87          & 3.27(1.58)          & 3.67(3.02)          & 94.18          & 3.35(1.51)          & 3.85(3.10)          & 76.45          & 6.73(4.23)          & 6.54(4.26)          & 76.91          & 7.99(4.38)          & 7.87(4.55)          & 65.46          & 3.21(2.51)          & 3.23(2.51)          & 65.52          & 3.11(2.51)          & 2.92(2.72)          \\
FL                                          & 94.82          & 3.42(1.07)          & 3.41(0.87)          & 93.59          & 3.27(1.12)          & 3.23(1.37)          & 76.41          & 2.83(1.66)          & 2.88(1.73)          & 76.12          & 3.10(2.58)          & 3.22(2.51)          & 63.08          & 2.03(2.03)          & 1.94(1.94)          & 64.02          & 2.18(2.18)          & 2.09(2.09)          \\
Mixup                                       & 94.76          & 2.86(1.37)          & 2.81(2.00)          & 95.50          & 6.87(1.18)          & 6.79(2.33)          & 78.47          & 8.68(2.14)          & 8.68(2.19)          & 78.74          & 8.92(3.69)          & 8.91(3.65)          & 65.81          & 1.92(1.92)          & 1.96(1.96)          & 66.41          & 2.41(1.97)          & 2.43(1.95)          \\
FLSD                                        & 94.77          & 3.86(0.83)          & 3.74(0.96)          & 93.26          & 3.92(0.93)          & 3.67(0.94)          & 76.20          & 2.86(2.86)          & 2.86(2.86)          & 76.61          & 3.29(2.04)          & 3.25(1.78)          & 63.56          & 1.93(1.93)          & 1.98(1.98)          & 64.02          & 1.85(1.85)          & 1.81(1.81)          \\
CRL                                         & 95.08          & 3.14(0.96)          & 3.11(1.25)          & 95.04          & 3.74(1.12)          & 3.73(2.03)          & 77.85          & 6.30(3.43)          & 6.26(3.56)          & 77.60          & 7.29(3.32)          & 7.14(3.31)          & 64.88          & 1.65(2.35)          & 1.52(2.34)          & 65.87          & 3.57(1.60)          & 3.56(1.52)          \\
CPC                                         & 95.04          & 5.05(1.89)          & 5.04(2.60)          & 95.36          & 4.78(1.52)          & 4.77(2.37)          & 77.23          & 13.29(3.74)         & 13.28(3.82)         & 77.50          & 13.32(2.96)         & 13.28(3.23)         & 65.70          & 3.41(3.41)          & 3.42(3.42)          & 66.44          & 3.93(3.93)          & 3.74(3.74)          \\
MbLS                                        & 95.25          & 1.16(1.16)          & 3.18(3.18)          & 95.13          & 1.38(1.38)          & 3.25(3.25)          & 77.92          & 4.01(4.01)          & 4.14(4.14)          & 77.45          & 5.49(5.49)          & 6.52(6.52)          & 64.74          & 1.64(1.64)          & 1.73(1.73)          & 65.81          & 1.62(1.62)          & 1.68(1.68)          \\
RegMixup                                    & 94.68          & 2.76(0.98)          & 2.67(0.92)          & 95.03          & 4.75(0.92)          & 4.74(0.94)          & 76.76          & 5.50(1.98)          & 5.48(1.98)          & 76.93          & 4.20(1.36)          & 4.15(1.92)          & 63.58          & 3.04(1.89)          & 3.04(1.81)          & 63.26          & 3.35(1.86)          & 3.32(1.68)          \\
AugMix  & 94.57 & 3.29(0.63) & 3.26(0.71) & 95.02 & 3.33(0.57) & 3.32(0.81) & 77.87 & 10.93(2.49) & 10.89(2.27) & 78.63 & 11.66(2.06) & 11.66(1.85) & 65.56 & 2.64(2.10) & 2.37(2.21) & 65.89 & 2.73(2.44) & 2.78(2.47) \\
FCL                                    & 95.44 & 0.76(0.76) & 0.75(0.75) & 95.25 & 0.95(0.95) & 1.25(1.25) & 78.19 & 3.71(2.14) & 3.74(2.11) & 78.98 & 4.19(3.23) & 4.88(3.27) &62.67 & 7.51(1.51) & 7.51(1.45) & 64.04 & 6.97(1.70) & 6.97(1.94) \\
RankMixup                                   & 94.88          & 2.59(0.57)          & 2.58(0.52)          & 94.25          & 3.24(0.65)          & 3.21(0.56)          & 77.11          & 3.46(1.49)          & 3.45(\textbf{1.42}) & 76.46          & 3.49(\textbf{1.10}) & 3.49(1.40)          & 64.97          & 1.49(1.49)          & 1.44(1.44)          & 64.89          & 1.57(1.57)          & 1.94(1.94)          \\
\rowcolor{gray!12}
CSM                                       & \textbf{95.79} & \textbf{0.54(0.54)} & \textbf{0.33(0.33)} & \textbf{95.80} & \textbf{0.54(0.54)} & \textbf{0.33(0.33)} & \textbf{78.84} & \textbf{1.29(1.29)} & \textbf{1.63}(1.63) & \textbf{79.17} & \textbf{1.46}(1.46) & \textbf{1.28(1.28)} & \textbf{66.99} & \textbf{1.29(1.29)} & \textbf{1.19(1.19)} & \textbf{68.20} & \textbf{1.33(1.33)} & \textbf{1.42(1.42)} \\ \bottomrule
\end{tabular}
}
\end{sc}
\end{center}
\vskip -0.1in
\end{table*}

\textbf{Balanced Learning}\quad 
Previous Mixup-based calibration methods adopt CE, Focal \cite{FLSD}, or certain ranking-based losses \cite{noh2023rankmixup} for calibration with Mixup samples. 
%We argue that they are not necessarily \textit{confidence-balanced} across mixup samples
We argue that they are not necessarily effective for augmented samples, especially in our case, where samples are \textit{required} to be confusingly similar via the conditional sampling and possibly almost indistinguishable in the feature or logit space. 
To reveal the deficiency of common loss objectives for soft-labeled data, we define confidence-balanced loss as follows.

\begin{definition} \label{def:CBL} Consider two soft-labeled data points associated with classes $i$ and $j$, with \textit{sharper} and \textit{softer} labels, respectively: $\widetilde{\bm q}_1 = \lambda_1 \bm q_i + (1 - \lambda_1) \bm q_{j}$ and $\widetilde{\bm q}_2 = \lambda_2 \bm q_i + (1 - \lambda_2) \bm q_{j}$, where $0.5 \le \lambda_2 < \lambda_1 < 1$.
A loss function $\mathcal{L}(\bm p, \bm q): \Delta^K \times \Delta^K \rightarrow \mathbb R$ is considered a confidence-balanced loss if and only if the optimal outputs $\widetilde{\bm p}^*_1, \widetilde{\bm p}^*_2$ for the following optimization problem:
\begin{align} 
&\min_{\widetilde{\bm p}_1, \widetilde{\bm p}_2}\mathcal{L}(\widetilde{\bm p}_1, \widetilde{\bm q}_1)+\mathcal{L}(\widetilde{\bm p}_2, \widetilde{\bm q}_2)\nonumber\\
&~~s.t. ~\|\widetilde{\bm p}_1-\widetilde{\bm p}_2\|_2^2 \le \delta 
\label{eq:problem} 
\end{align} 
satisfy the condition that the balance function $\beta(\widetilde{\bm p}_1, \widetilde{\bm p}_2) = \|\widetilde{\bm p}_1-\widetilde{\bm q}_1\|_2^2 - \|\widetilde{\bm p}_2-\widetilde{\bm q}_2\|_2^2$ equals zero for all $\delta \ge 0$ and for all possible data pairs. %Here, $\Delta^K$ represents the $K$-dimensional probability simplex space. 
The prediction proximity condition $\|\widetilde{\bm p}_1-\widetilde{\bm p}_2\|_2^2 < \delta$ ensures the non-discriminability of correlated Mixup samples.
\end{definition}
Intuitively, $\|\widetilde{\bm p}_1-\widetilde{\bm q}_1\|_2^2$ and $\|\widetilde{\bm p}_2-\widetilde{\bm q}_2\|_2^2$ estimate how well the model fits the two soft labels, respectively. With $\beta(\bm p_1, \bm p_2) > 0$, $\|\widetilde{\bm p}_2-\widetilde{\bm q}_2\|_2^2$ is smaller and the model fits the softer label $\widetilde{\bm q}_2$ more effectively. In contrast, when $\beta(\widetilde{\bm p}_1, \widetilde{\bm p}_2) < 0$, $\|\widetilde{\bm p}_1-\widetilde{\bm q}_1\|_2^2$ is smaller and the model adheres more closely to the sharper label $\widetilde{\bm q}_1$. The network learns without such a bias only when $\beta(\widetilde{\bm p}_1, \widetilde{\bm p}_2) = 0$. These cases are also intuitively illustrated in \cref{fig:framework}.

It's worth noting that the majority of common loss functions are \textit{not} confidence-balanced. For example, ranking- or margin-based losses presume poor reliability of Mixup labels. Consequently, their learning behavior is basically blind to the exact class posteriors, therefore difficult to establish such balances. For CE and Focal losses, their patterns are both empirically and theoretically predictable. Denoting the optimal $\bm p_a$ as $\bm p^{CE}_a$ and $\bm p^{FL}_a$ ($a\in \{1,2\}$) for CE and Focal losses, respectively, we can determine the sign of score $\beta$:
\begin{proposition}
\label{prop:CE_prop}
For $\forall \delta \ge \|\widetilde{\bm q}_1-\widetilde{\bm q}_2\|_2^2$, we have $\beta_{\delta}=\beta(\bm p^{CE}_1, \bm p^{CE}_2) = 0$, while for $\forall \delta < \|\widetilde{\bm q}_1-\widetilde{\bm q}_2\|_2^2$, we have $\beta_{\delta}=\beta(\bm p^{CE}_1, \bm p^{CE}_2) <0$.
\end{proposition}
It can be seen that similar augmented pairs can exhibit negative scores, indicating shifted confidence toward sharper class distributions. In contrast, the behavior of the Focal loss is almost the opposite in such cases, producing positive $\beta$ scores for similar samples:
\begin{proposition}
\label{prop:FL_prop}
For $\gamma_{FL}=1$ and $\forall \delta < \|\widetilde{\bm q}_1-\widetilde{\bm q}_2\|_2^2$, we have $\beta_{\delta}=\beta(\bm p^{FL}_1, \bm p^{FL}_2) > 0$.
\end{proposition}
This imbalanced learning severely degrades the calibration improvement from Mixup samples by dragging the overall confidence towards certain sides. Aiming at obtaining balanced confidence, we present the following proposition:
\begin{proposition}
\label{prop:L2_prop}
$\forall \delta \ge 0$, we have $\beta(\bm p^{L2}_1, \bm p^{L2}_2) = 0$,
\end{proposition}
where $\bm p^{L2}_a$ is the optimal solution adopting the L2 loss:
\begin{align}
\mathcal{L}_{L2}(\bm p, \bm q) = \frac{1}{K}\|\bm p-\bm q\|_2^2.
\label{eq:L2_loss}
\end{align}
The formal proofs of Propositions \ref{prop:CE_prop}, \ref{prop:FL_prop}, and \ref{prop:L2_prop} are provided in Appendix \ref{sec:proofs_loss}. The proposition reveals that when two similar samples exceed the model's discriminability, $\mathcal{L}_2$ loss tends to balance the learned labels of both the harder and softer instances, instead of tending to fit a specific one of them.
Notably, this objective is tailored for learning on soft-labeled data, in contrast to the calibration of one-hot-labeled samples explored in \cite{liang2024calibrating}. 
In this way, we can formulate our overall objective as a simple combination of the CE and L2 losses, \textit{i.e.},
\begin{align}
\mathcal{L}_{overall}=\sum_{i=1}^{N}\mathcal{L}_{CE}(\bm p_i, y_i) + \sum_{a=1}^{A}\mathcal{L}_{L2}(\widetilde{\bm p}_a, \widetilde{\bm q}_a).
\label{eq:overall_loss}
\end{align}

\section{Experiment}
\label{sec:exp}

\begin{table}[!t]
\begin{center}
\caption{\textbf{Comparison on the \textsc{Eq-Data} setting} on CIFAR-10 with methods that use non-augmented data. Results in parentheses are post-temperature results.} \label{table_eq_data}
\vskip 0.1in
\begin{center}
\begin{small}
\begin{sc}
\resizebox{0.48\textwidth}{!}{
\begin{tabular}{lccccccc}
\toprule
Metric           & LS         & FL         & FLSD       & MbLS       & FCL        & \textbf{CSM (Ours)}        \\ 
\midrule
ACC$\uparrow$    & 94.87      & 94.82      & 94.77      & 95.25      & \textbf{95.44}      & 95.02      \\
ECE$\downarrow$  & 3.27(1.58) & 3.42(1.07) & 3.86(0.83) & 1.16(1.16) & \textbf{0.76}(0.76) & 0.91(\textbf{0.41}) \\
AECE$\downarrow$ & 3.67(3.02) & 3.41(0.87) & 3.74(0.96) & 3.18(3.18) & 0.75(0.75) & \textbf{0.68}(\textbf{0.42}) \\ 
\bottomrule
\end{tabular}
}
\end{sc}
\end{small}
\end{center}
\vskip -0.1in
\end{center}
\end{table}

We conduct experiments with DNNs of different architectures, including ResNet-50/101 \cite{resnet}, Wide-ResNet-26-10 \cite{zagoruyko2016wide}, and DenseNet-121 \cite{huang2017densely}.
We adopt CIFAR-10, CIFAR-100 \cite{cifar}, and Tiny-ImageNet \cite{tiny} for calibration performance and out-of-distribution (OOD) robustness comparisons. 
For augmentation, we adopt the class-conditional elucidating diffusion model (EDM) \cite{karras2022elucidating} to generate soft-labeled samples. 
We select the standard ECE \cite{ECE} and AECE \cite{AECE} protocols as our primary evaluation metrics for calibration, while following the widely-used area under the receiver operating characteristics (AUROC) protocol \cite{liu2020simple, pinto2022using} to evaluate the OOD detection performance.
Detailed descriptions of our benchmark datasets, evaluation protocols, and compared methods are presented in \cref{sec:more_exp_info}.

\begin{comment}
\textbf{Compared Baselines~} 
We adopt diverse training-time methods for comparison including the vanilla CE loss. Specifically, we compare calibration performance with the traditional regularization-based methods (ECP \cite{pereyra2017regularizing}, MMCE \cite{kumar2018trainable}, LS \cite{muller2019does}, mbLS \cite{liu2022devil}, FL \cite{focal_loss}, FLSD \cite{FLSD}, CPC \cite{cheng2022calibrating}, FCL \cite{liang2024calibrating}) and data-driven methods (Mixup \cite{zhang2018mixup}, RegMixup \cite{pinto2022using}, AugMix \cite{hendrycks2019augmix}, RankMixup \cite{noh2023rankmixup}).
\end{comment}

\textbf{Training Time}\quad During training, data-aware calibration methods supplement each example with additional augmented samples. To control the training time, we limit the maximum number of auxiliary data associated with each dataset instance as $N_{\text{aug}}=3$ considering the settings in \cite{noh2023rankmixup}. In fact, our method adopts $N_{\text{aug}}=2$ for CIFAR-10/100 and $N_{\text{aug}}=1$ for Tiny-ImageNet, displaying strong data efficiency. To make a fully fair comparison with non-augmentation methods, we also show results with a different data setting \textsc{Eq-Data}: During training, we limit the number of iterations per epoch to be $\lceil\frac{\text{dataset\_size}}{(1+N_{\text{aug}})(\text{batch\_size})}\rceil$, which ensures the total amount of training data per epoch to be consistent across methods.

\textbf{Training Details}\quad We use our proposed method to generate augmented samples for each dataset based on the code and checkpoints from \cite{edm} and \cite{dm_improves_AT}. We generate $159,840$; $158,400$; and $318,400$ CSM-augmented samples for CIFAR-10, CIFAR-100, and Tiny-ImageNet, respectively. The setup for training and testing in our work follows \cite{noh2023rankmixup}, while we also implement our approach based on their public code. Specifically, we conduct 200-epoch training on CIFAR-10 and CIFAR-100 and 100-epoch training on Tiny-ImageNet using the SGD optimizer of momentum set to $0.9$. We adopt multi-step learning rate schedule which decreases from 0.1 to 0.01 and 0.001 at epochs 81 and 121 for CIFAR-10/100, or epochs 40 and 60 for Tiny-ImageNet, respectively. The weight decay is set to $5\times10^{-4}$. We select the scaling hyperparameter $s$ as $4.0$ for CIFAR-10/Tiny-ImageNet and $2.3$ for CIFAR-100 using their validation sets.

\subsection{Results}
\textbf{State-of-the-Art Comparison}\quad We provide comprehensive comparisons with state-of-the-art approaches in \cref{table_main}. From the results, we can derive the following key observations: 
(1) Our CSM outperforms traditional regularization-based methods (\eg, Label Smoothing and FLSD) by large margins via confidence-aware data augmentation, indicating the strong effectiveness of our method to leverage confidence knowledge embedded in the learning samples. 
(2) CSM gives comparable or superior results for ECE and ACEC consistently across all compared benchmarks including CIFAR-10, CIFAR-100, and Tiny-ImageNet, especially for the pre-temperature-scaling results. Such performance suggests that the confidence learning driven by data is crucial for model calibration, and our calibration-aware augmentation scheme can reliably generate the required sample-label pairs. 
(3) Our method also outperforms Mixup-based calibration methods, including the state-of-the-art RankMixup, suggesting that our generated augmentations enable better sample-confidence alignment compared to traditional Mixup outputs in terms of both fidelity and coherence.
(4) With the post-hoc temperature scaling \cite{ECE}, optimal temperatures in our method are mostly determined as $1.0$ while compared methods generally exhibit larger $T$ values, further validating that CSM effectively solves the over-confidence issue embedded in ordinary one-hot datasets and the compared learning paradigms.

We also share results on larger datasets or architectures by comparing CSM with representative methods on ImageNet with the ResNet-50 and Swin-Transformer architectures in \cref{table_imagenet_r50} and \cref{table_imagenet_swin}, respectively. Our method performs equally or more effectively compared to these methods, especially to the mixup-based methods. 

\begin{table}[!t]
\begin{center}
\caption{\textbf{Comparison on ImageNet using ResNet-50 architecture.}}
\label{table_imagenet_r50}
\vskip 0.1in
\begin{center}
\begin{small}
\begin{sc}
\resizebox{0.48\textwidth}{!}{
\begin{tabular}{lccccccc}
\hline
Metrics          & CE    & Mixup & MbLS  & RegMixup & RankMixup & CALS          & CSM (Ours)            \\ \hline
ACC              & 73.96 & 75.84 & 75.39 & 75.64    & 74.86     & 76.44         & \textbf{79.87} \\
ECE$\downarrow$  & 9.10  & 7.07  & 4.07  & 5.34     & 3.93      & 1.46          & \textbf{1.32}  \\
AECE$\downarrow$ & 9.24  & 7.09  & 4.14  & 5.42     & 3.92      & \textbf{1.32} & 1.35           \\ \hline
\end{tabular}
}
\end{sc}
\end{small}
\end{center}
\vskip -0.1in
\end{center}
\end{table}

\begin{table}[!t]
\begin{center}
\caption{\textbf{Comparison on ImageNet using Swin-Transformer-V2 architecture.}}
\label{table_imagenet_swin}
\vskip 0.1in
\begin{center}
\begin{small}
\begin{sc}
\resizebox{0.48\textwidth}{!}{
\begin{tabular}{lccccccc}
\hline
Metrics          & CE    & LS    & FL    & FLSD  & MbLS  & CALS          & CSM (Ours)           \\ \hline
ACC              & 75.60 & 75.42 & 75.60 & 74.70 & 77.18 & 77.10         & \textbf{81.08} \\
ECE$\downarrow$  & 9.95  & 7.32  & 3.19  & 2.44  & 1.95  & 1.61          & \textbf{1.49}  \\
AECE$\downarrow$ & 9.94  & 7.33  & 3.18  & 2.37  & 1.73  & \textbf{1.69} & 1.86           \\ \hline
\end{tabular}
}
\end{sc}
\end{small}
\end{center}
\vskip -0.1in
\end{center}
\end{table}

\begin{comment}
\cref{table_wide_resnet} and \cref{table_densenet} (in Appendix) also provides qualitative comparisons using Wide-ResNet-26-10 \cite{zagoruyko2016wide} and DenseNet-121 \cite{huang2017densely} as the network, respectively. The results demonstrate similar property to those in \cref{table_main}, indicating that our method performs consistently well across various architectures. Furthermore, our CSM displays even greater ECE margin from the traditional mixup-based methods \cite{zhang2018mixup, pinto2022using, noh2023rankmixup}, verifying its successful generation and learning of authentic confidence-aware samples. Meanwhile, we provide results on the \textsc{Eq-Data} setting in \cref{table_eq_data} for comparisons with non-augmentation methods regarding training time. It is observed that our CSM can still achieve competitive or superior performance. Furthermore, the post-temperature ECE and AECE can remain stable regardless of the number of training data in an epoch, verifying the stability of CSM learning scheme.
\end{comment}
We also provide qualitative comparisons using Wide-ResNet-26-10 \cite{zagoruyko2016wide} and DenseNet-121 \cite{huang2017densely} as the network in \cref{sec:more_exp_results}, which demonstrate similar properties to those in \cref{table_main}. Meanwhile, we provide results on the \textsc{Eq-Data} setting in \cref{table_eq_data} for comparisons with non-augmentation methods for equalized training time. It is observed that our CSM can still achieve competitive or superior performance, particularly in consistently superior post-temperature ECE and AECE. Comparisons of the estimated training time is provided in \cref{sec:more_exp_results}, where one can observe that the number of augmented samples per batch is the major factor affecting training time.

\textbf{Out-of-Distribution Detection}\quad To validate the network calibration from different perspectives, we conduct qualitative comparisons for the out-of-distribution (OOD) detection with the proposed CSM. We measure the sample uncertainty by the entropy of the softmax output $\bm p_i = \operatorname{softmax}(\bm f_{\theta}(\bm x_i))$. Therefore, we report AUROC scores comparisons in \cref{table_ood}. In the table, our proposed CSM achieves superior and comparable results on the CIFAR-10 and CIFAR-100 datasets, respectively. Such results indicate that our approach learns precise sample confidence levels and is therefore also superior in detecting distributional shifts, verifying the significance of explicit soft-labeled samples in OOD detection.

\begin{table}[!t]
\begin{center}
\caption{\textbf{AUROC (\%) for robustness evaluation under distribution shifts.} Higher values indicate better performance. We use CIFAR-10 and CIFAR-100 as in-distribution datasets, with each serving as the OOD dataset for the other, alongside an external Tiny-ImageNet dataset.} \label{table_ood}
\vskip 0.1in
\begin{center}
\begin{small}
\begin{sc}
\resizebox{0.46\textwidth}{!}{
\begin{tabular}{lcccc}
\toprule
ID dataset  & \multicolumn{2}{c}{CIFAR-10}     & \multicolumn{2}{c}{CIFAR-100}    \\
OOD dataset & CIFAR-100       & Tiny-ImageNet  & CIFAR-10        & Tiny-ImageNet  \\ \midrule
TS           & 86.73          & 88.06          & 76.38          & 80.12          \\
LS           & 75.23          & 76.29          & 72.92          & 77.54          \\
FL           & 85.63          & 86.84          & 78.37          & 81.23          \\
Mixup        & 82.07          & 84.09          & 78.46          & \textbf{81.55} \\
CPC          & 85.28          & 85.15          & 74.68          & 76.66          \\
MbLS         & 86.20          & 87.55          & 73.88          & 79.19          \\
RegMixup     & 87.82          & 87.54          & 76.00          & 80.72          \\
RankMixup    & 87.82          & 88.94          & 78.64          & 80.67          \\
\rowcolor{gray!12}
CSM         & \textbf{91.82} & \textbf{91.05} & \textbf{79.02} & 80.45          \\ \bottomrule
\end{tabular}
}
\end{sc}
\end{small}
\end{center}
\vskip -0.1in
\end{center}
\end{table}

\begin{comment}
\begin{figure}[ht]
\vskip 0.2in
\begin{center}
\centerline{
\includegraphics[width=0.75\columnwidth]{figures/percentage_samples.pdf}
}
\caption{Classification ACC, ECE, and AECE on CIFAR-100 \textit{w.r.t} different number of augmented samples relative to the dataset size.}
\label{percentage_samples}
\end{center}
\vskip -0.2in
\end{figure}
\end{comment}

\begin{comment}
\begin{figure*}[ht]
\vskip 0.2in
\begin{center}
\centerline{
\includegraphics[height=0.26\textwidth]{figures/scaling_param_oe_ue.pdf}
\includegraphics[height=0.26\textwidth]{figures/scaling_param_ece_aece.pdf}
\includegraphics[height=0.26\textwidth]{figures/percentage_samples.pdf}
}
\caption{Analysis of the amount of augmented data and hyperparameter $s\in [2.0, 6.0]$. (a) OE and UE \textit{w.r.t.} different $s$ values. (b) ECE/AECE and their post-temperature values \textit{w.r.t} different $s$ values. (c) Classification ACC, ECE, and AECE on CIFAR-100 \textit{w.r.t} different number of augmented samples relative to the dataset size.}
\label{parameters_nsample_and_s}
\end{center}
\vskip -0.2in
\end{figure*}
\end{comment}

\subsection{Discussions}
We discuss the properties of our method with extensive experiments below. More results are available in \cref{sec:more_exp_results}.

\begin{table}[!t]
\begin{center}
\caption{\textbf{Ablation Study.} Variants regarding augmentations, annotations, and learning objectives are compared. Results are from CIFAR-100 with ResNet-50.} \label{table_ablations}
\vskip 0.1in
\begin{center}
\begin{small}
\begin{sc}
\resizebox{0.48\textwidth}{!}{
%\begin{tabular}{c|c|ccc|ccc|ccc|ccc}
%\begin{tabular}{llcccccccccccc}
\begin{tabular}{lcccccc}
\toprule
Variants & ACC$\uparrow$  & $\text{ECE}\downarrow$  & $\text{AECE}\downarrow$   & $\text{ECE}^{\text{PT}}\downarrow$       & $\text{ECE}^{\text{PT}}\downarrow$      & $T$   \\ 
\midrule
1. w/o Augmented Data                         & 77.48          & 12.94         & 12.94         & 2.63          & 2.66          & 1.6 \\
2. CSM (L2) w/o Reanno.     & 78.92          & 5.20          & 5.20          & 3.33         & 3.24          & 1.2 \\
3. CSM (L2) w/ CLIP labels. & 66.60          & 52.78         & 52.78         &  -           &  -            & - \\
4. CSM (L2) w/ One-hot Aug.      & \textbf{79.24}          & 10.84         &  10.84        &  2.48        &  2.41         & 1.5 \\
\midrule
5. Mixup (CE)                 & 78.47          & 8.68          & 8.68          & 2.14          & 2.19          & 1.3 \\
6. Mixup (FL)                 & 79.64          & 2.45          & 2.44          & 2.45          & 2.44          & 1.0   \\
7. Mixup (L2)                 & 78.88          & 2.49          & 2.62          & 2.49          & 2.62          & 1.0   \\
\midrule
8. CSM (CE)                 & 78.93 & 2.47          & 2.14         & 2.47          & 2.14          & 1.0   \\
9. CSM (FL)                 & 77.65          & 2.43          & 2.30          & 2.16         & 2.09          & 0.9 \\
\rowcolor{gray!12}
10. CSM (L2, ours)                 & 78.84          & \textbf{1.29} & \textbf{1.63} & \textbf{1.29} & \textbf{1.63} & 1.0   \\
\bottomrule
\end{tabular}
}
\end{sc}
\end{small}
\end{center}
\vskip -0.1in
\end{center}
\end{table}

\begin{figure*}[ht]
\vskip 0.2in
\begin{center}
\centerline{
\includegraphics[width=0.24\textwidth]{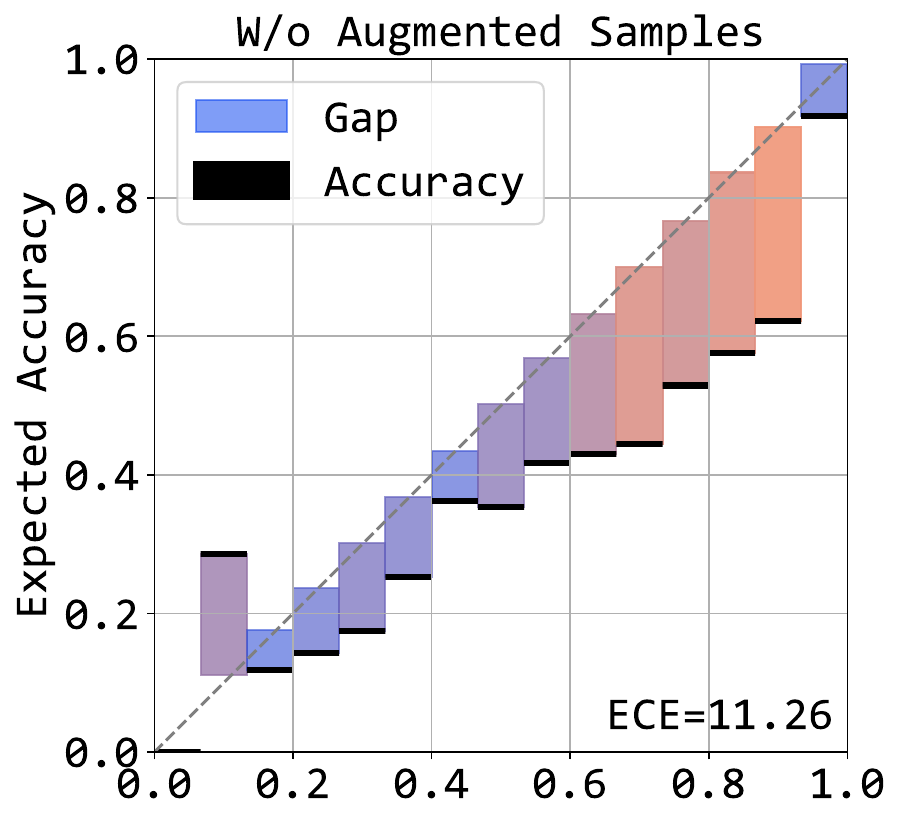}
\includegraphics[width=0.24\textwidth]{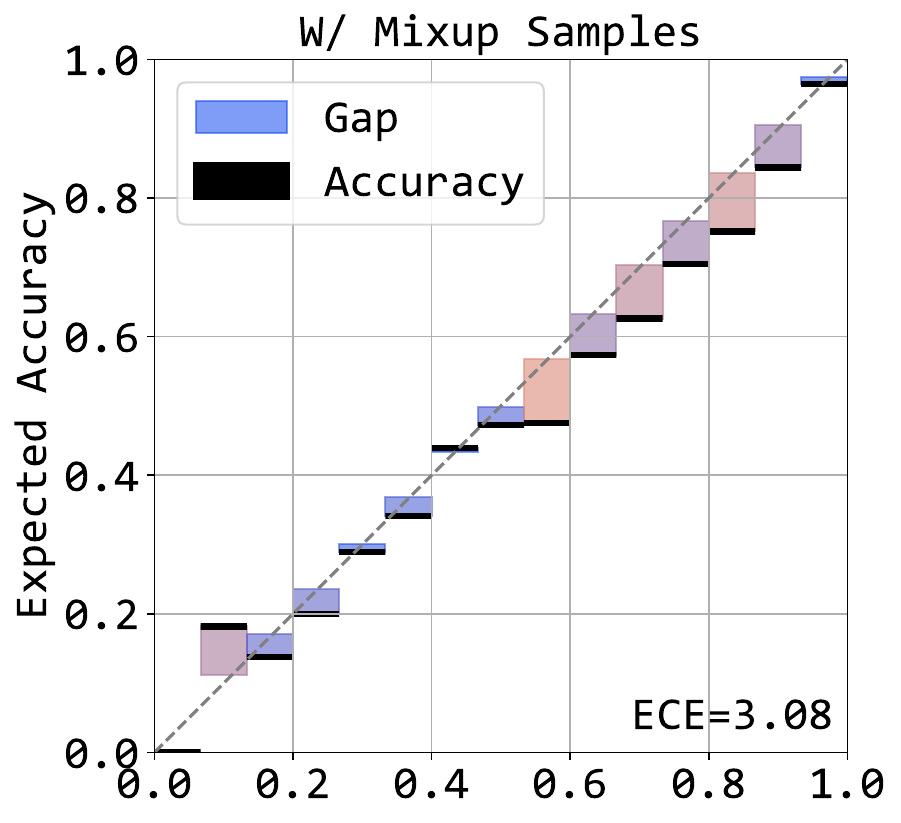}
\includegraphics[width=0.24\textwidth]{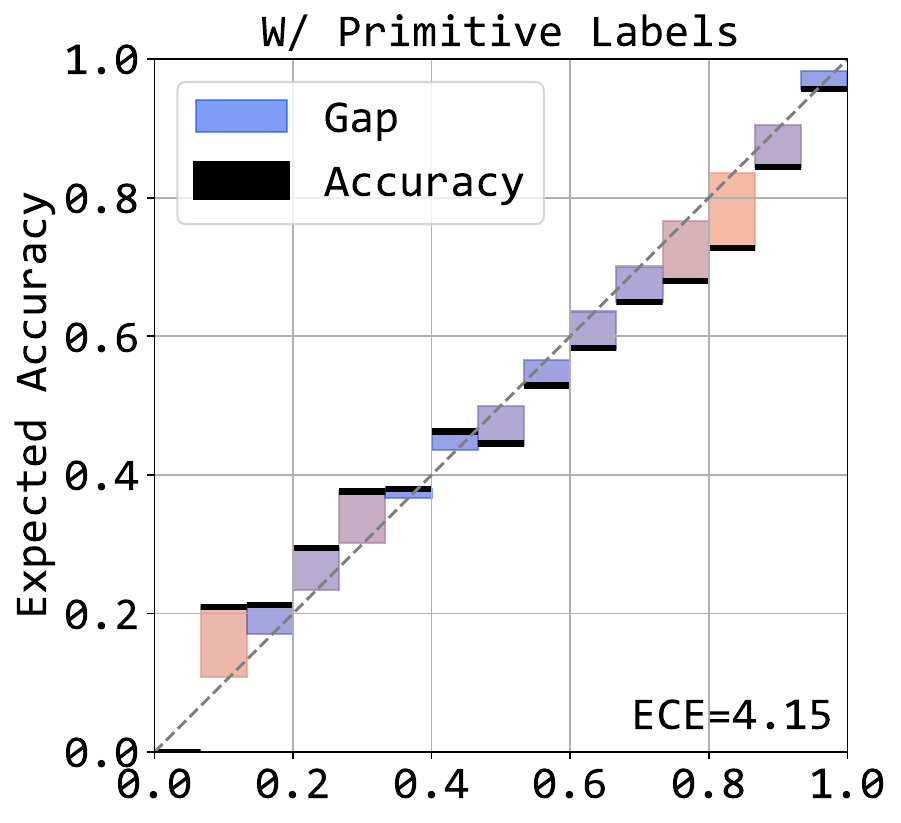}
\includegraphics[width=0.24\textwidth]{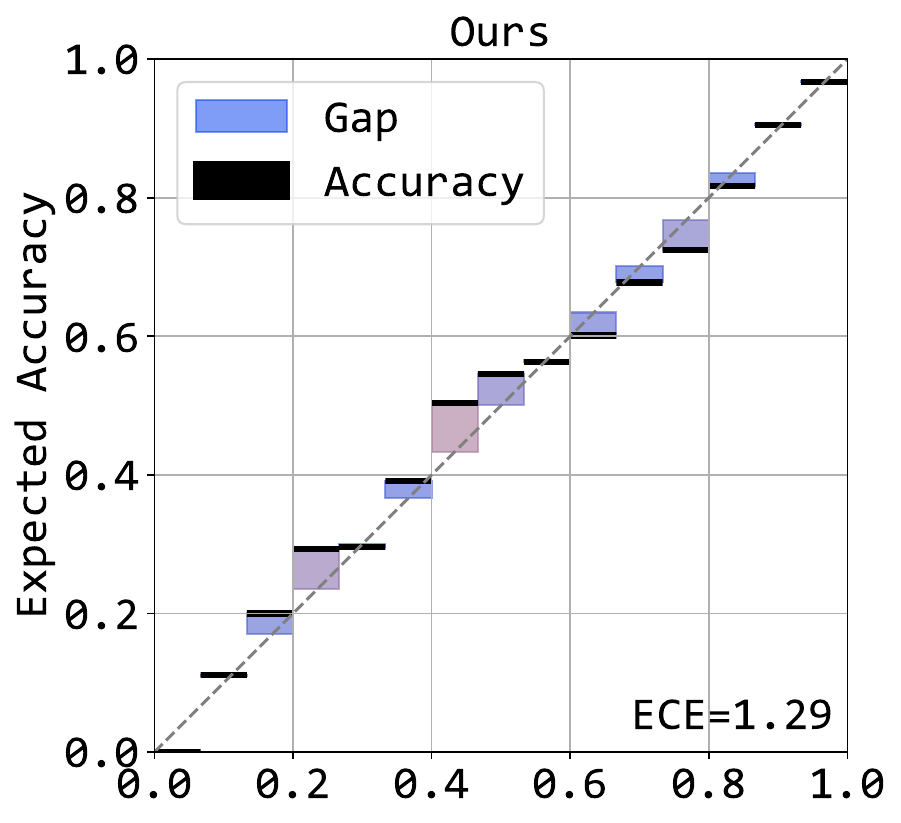}
}
\caption{\textbf{Reliability diagram} of various methods before temperature scaling.}
\label{reliability_diagrams}
\end{center}
\vskip -0.2in
\end{figure*}

\begin{figure}[t]
\vskip 0.2in
\begin{center}
\centerline{
\includegraphics[width=0.49\textwidth]{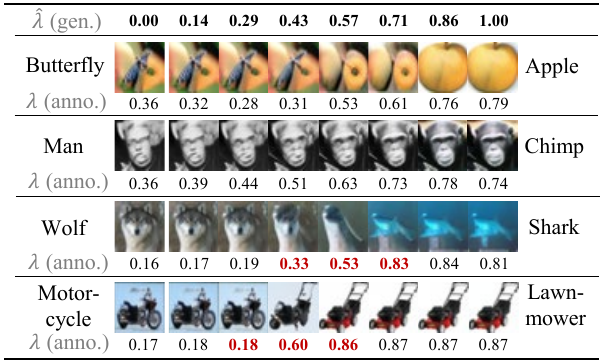}
}
\caption{\textbf{Visualization of generated samples}, including augmented sample sets, mixing coefficient $\hat{\lambda}$ during generation (gen.), and their reannotated $\lambda$ (anno.) for soft labels.}
\label{fig:visualization_sample_lambda}
\end{center}
\vskip -0.2in
\end{figure}

\textbf{Ablation Study}\quad We compare our CSM with variants in \cref{table_ablations}, testing different augmentation types, soft labels, and loss functions. We can observe that Mixup and CSM variants outperform the standard CE loss, improving both ECE and accuracy, highlighting the importance of data-driven network calibration. 
Compared with the variant w/o reannotation (or using the generation labels, Var. 2), our reannotated variant produces significantly better calibration performances, validating the accurate confidence estimation in our refined labels.
Meanwhile, using the vanilla CLIP annotations (Var. 3) yields the worst ACC and pre-temperature calibration errors, primarily due to the noisy information by annotating all classes. 
Directly adopting class-conditioned augmentations from the diffusion model (Var. 4) can slightly rise the prediction accuracy, as also evidenced by the augmented classification literature. However, it fails to improve model calibration due to the lack of soft-labeled samples.

Among the losses used in CSM (Var. 8-10), the L2 loss is the only one that reduces ECE, AECE, and post-TS results below 2.0, suggesting it effectively balances learned confidence in augmentations. In contrast, CE and FL losses often require temperature adjustments (Var. 5,9), with CE favoring sharper labels and FL for softer ones, aligning with our theoretical expectations from \cref{sec:CRL}.
CSM variants, particularly with CE and L2 losses, significantly outperform traditional Mixup in terms of calibration, verifying the effectiveness of CSM’s realistic in-domain augmentations. 
%However, applying L2 loss to traditional Mixup samples doesn’t produce significantly better results, as Mixup samples are independently sampled, which makes the learning of sample-label correspondence much easier for the model. In contrast, CSM uses conditioned dual-sampling of one-hot samples $\widetilde{\bm x}^{0}$ and $\widetilde{\bm x}^{1}$, leading to difficult fitting yet better calibration. 

However, applying L2 loss to the traditional Mixup samples does not yield significantly better results (Var. 5-7). It is noteworthy that CSM uses conditioned dual-sampling of one-hot samples $\widetilde{\bm x}^{0}$ and $\widetilde{\bm x}^{1}$, which are correlated considering joint latent variable $\bm x^{(T)}$, while every pair of them in Mixup is sampled independently, which makes the soft-labeled samples much easier to be distinguished apart by the model. Therefore, traditional Mixup with L2 loss cannot perform noticeably better on all metrics. 

\textbf{Augmented Samples and Annotations}\quad To analyze the properties of augmented samples, we visualize some case images in Fig. \ref{fig:visualization_sample_lambda}. It can be seen that CSM can produce semantically sound intermediate samples compared to Mixup's simple overlap. 
Meanwhile, these samples exhibit characteristics of both classes with gradual semantic change from one category to another. 
For images with abrupt transitions (in red numbers) or inaccurate class posteriors, our annotation paradigm yields more precise $\lambda$ values compared to the generation phase $\hat{\lambda}$.

\textbf{Confidence Value Adjustment}\quad To analyze the learned confidence values, we compare the reliability diagrams of different methods across \cref{reliability_diagrams}. We examine CSM alongside three variants: 1) without augmentations, 2) with vanilla Mixup, and 3) using labels from generation. The augmentation-free variant shows severe over-confidence across the middle-to-high confidence range, while the Mixup variant also exhibits over-confidence, though less severely. Regularization-based methods that calibrate with one-hot labels can show similar over-confidence, highlighting common issues in confidence estimation. CSM with primitive labels $\hat{\lambda}$, however, is both over- and under-confident at higher and lower ranges, respectively, suggesting a misalignment between annotations and confidence levels. In contrast, CSM demonstrates precise confidence estimation across both low and high ranges, with only slight errors in the middle, confirming the effectiveness of our confidence-aware augmentation.

\begin{figure*}[t]
\vskip 0.2in
\centering
\subfigure[]{
    \includegraphics[width=0.31\linewidth]{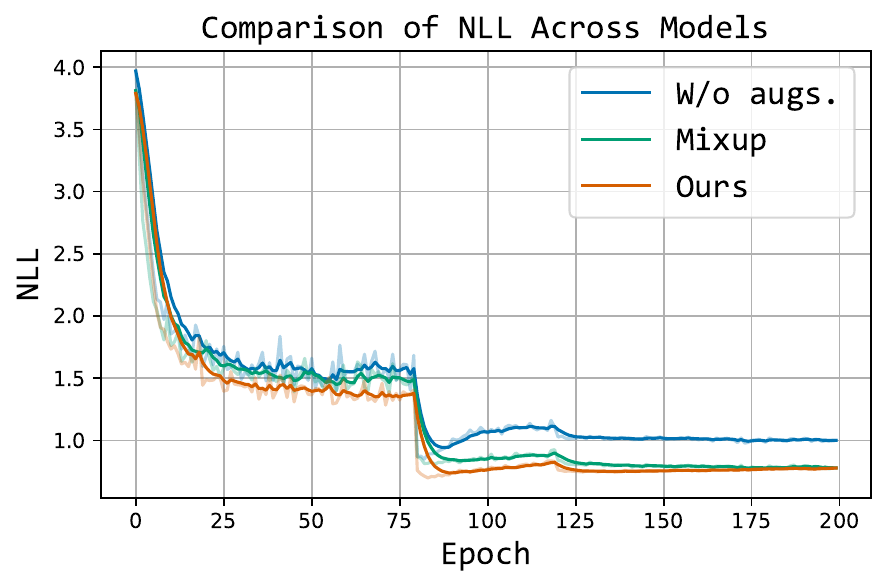}
}
\subfigure[]{
    \includegraphics[width=0.31\linewidth]{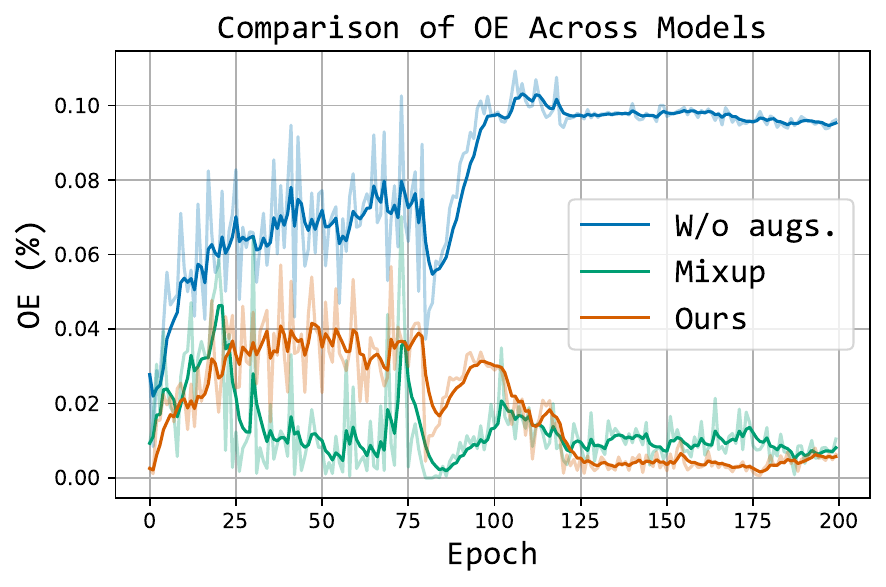}
}
\subfigure[]{
    \includegraphics[width=0.31\linewidth]{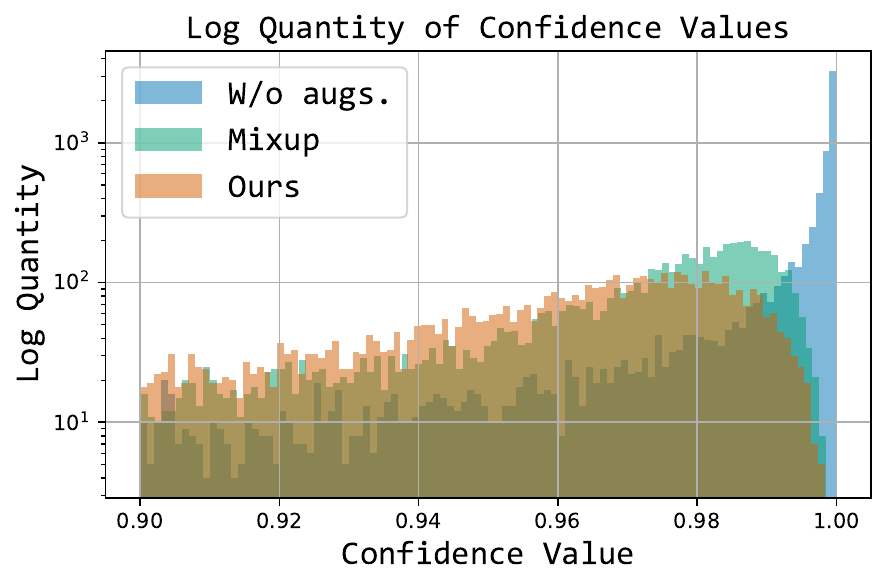}
}
\caption{\textbf{Training Characteristics.} (a) NLL values of different methods on CIFAR-100 validation split across training epochs. (b) Over-confidence Errors (OE) of different methods on CIFAR-100 validation split across training epochs. (c) Log-quantity of top confidence values in CIFAR-100 test split among different methods.}
\label{calib_in_training_and_conf_vals}
\vskip -0.2in
\end{figure*}

We also examine the distribution of top confidence values in \cref{calib_in_training_and_conf_vals}(c). Without soft labels, predicted confidence peaks at 1.00, while Mixup-augmented training peaks at around 0.99. Our CSM produces a more even distribution, effectively reducing the over-confidence observed in calibration methods using one-hot labels.

\textbf{Calibration through Training}\quad
To dynamically study the training-time behavior of CSM, we plot the Negative Log-Likelihood (NLL) and Overconfidence Errors (OE) in \cref{calib_in_training_and_conf_vals}(a)-(b) using exponential moving averages for better visualization. From the plot, we can observe that better classification has already been achieved in early epochs while the strong calibration is not evident until the second adjustment of the learning rate. This is mainly due to the order of network fitting, which prioritizes easy one-hot samples but learns the actual confidence in hard soft labels later. The lower calibration error of Mixup during early epochs also indicates that the larger difficulty in fitting confidence of our augmentations compared to Mixup samples. However, CSM can eventually achieve lower calibration error with a fine-grained learning rate for better fitting, validating the effectiveness of our proposed learning scheme.

\begin{figure}[t]
\vskip 0.2in
\begin{center}
\centerline{
\subfigure[]{
    \includegraphics[height=0.22\textwidth]{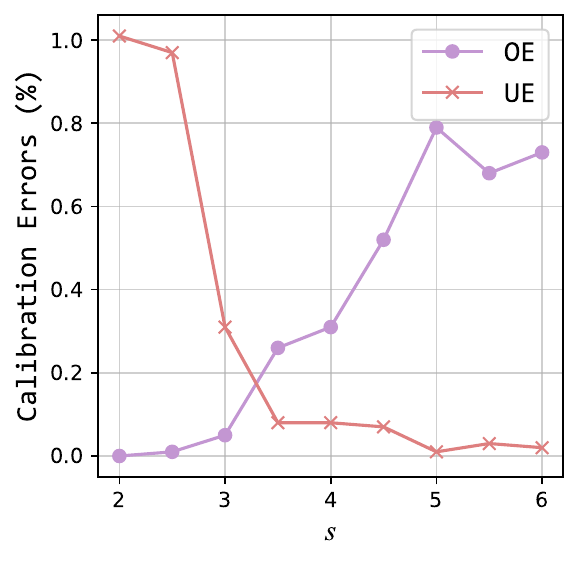}
}
\subfigure[]{
    \includegraphics[height=0.22\textwidth]{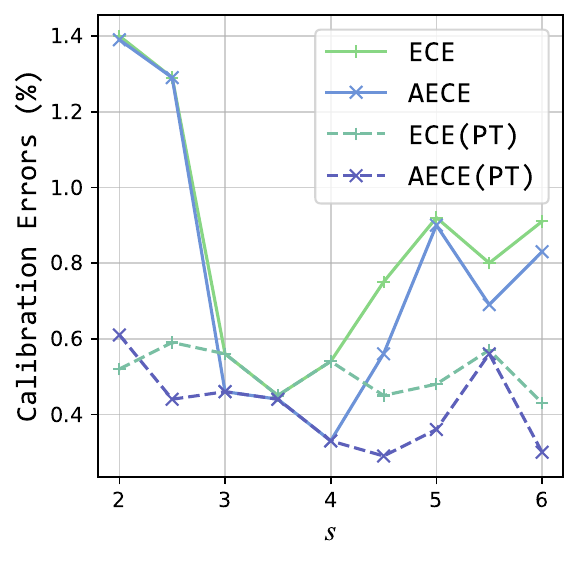}
}
%\subfigure[]{
%    \includegraphics[height=0.22\textwidth]{figures/percentage_samples.pdf}
%}
}
\caption{\textbf{Analysis of the hyperparameter} $s\in [2.0, 6.0]$. (a) OE and UE \textit{w.r.t.} different $s$ values. (b) ECE/AECE and their post-temperature values \textit{w.r.t} different $s$ values.}
\label{parameters_s}
\end{center}
\vskip -0.2in
\end{figure}

\textbf{Scaling Factor}\quad To analyze the influence of hyperparameter $s$, we plot various calibration results on CIFAR-10 in \cref{parameters_s}, including Overconfidence Errors (OE) and Underconfidence Errors (UE) \cite{thulasidasan2019mixup}. Due to the scaling effect of $s$ values to the confidence levels, the label distribution can shift towards shaper and softer ones with higher and lower $s$, respectively. This is evident in \cref{parameters_s}(a), where OE increases while UE decreases when the labels are shifted towards traditional one-hot annotations by large $s$, meaning that the model behaves over-confident about predictions. Consequently, ECE and AECE values increase a lot when $s$ leaves a certain range as shown in \cref{parameters_s}(b). Nevertheless, with post-hoc temperature scaling, the calibration errors can become consistently low regardless of $s$ values, verifying the stability of the proposed CSM.

\begin{table}[!t]
\begin{center}
\caption{\textbf{Integration with Test-Time Calibration Methods.}} \label{table_test_time}
\vskip 0.1in
\begin{center}
\begin{small}
\begin{sc}
\resizebox{0.48\textwidth}{!}{
%\begin{tabular}{c|c|ccc|ccc|ccc|ccc}
%\begin{tabular}{llcccccccccccc}
\begin{tabular}{lccc}
\toprule
Variants & ECE$\downarrow$  & AECE$\downarrow$  & PIECE \cite{xiong2023proximity}$\downarrow$ \\ 
\midrule
CSM          & \textbf{1.29} & 1.63 & 3.16 \\
CSM + TTA     & 1.39 & \textbf{1.53} & 3.15 \\
CSM + ProCal  & 1.89 & 1.82 & \textbf{3.11} \\
\bottomrule
\end{tabular}
}
\end{sc}
\end{small}
\end{center}
\vskip -0.1in
\end{center}
\end{table}

\textbf{Integrating Test-time Methods}\quad While CSM introduces training-time augmentations for softened labels, there exists test-time augmentation methods targeting the same goal. Meanwhile, proximity problems have also been investigated in inference time. 
We conduct experiments to combine these methods with our CSM, as demonstrated in \cref{table_test_time}. Integrated with test-time augmentations (TTA), the method balances ECE and AECE effectively, achieving an optimized AECE of 1.53 on CIFAR-100. Compared to \cite{hekler2023test} using test-time sample-wise scaling, CSM employs training-time augmentation with inter-sample augmentations to expand the proximity space, enhancing calibration robustness. Combined with proximity method ProCal, we find that the proximity-informed metric PIECE displays better results, which validates the robustness growth related to proximity from the integration. Despite these growths, the overall calibration improvement is relatively small, indicating the effective sample augmentation and debiased proximity learning of our method.

\section{Conclusion}
\label{sec:conclusion}
In conclusion, CSM is a novel framework designed to bridge the optimization gap between the calibration objective and the data of full certainty. By generating augmented samples with semantic mixing and reannotating them with confidence scores via diffusion models, CSM enables a more accurate alignment between model predictions and their true likelihoods. Our exploration of balanced loss functions further enhances the new data representation paradigm, enhancing it an integrated pipeline for superior model calibration. Theoretical and practical evidence validate the effectiveness of CSM for strong sample-label augmentations. Meanwhile, various ablation results demonstrate CSM's balanced learning of true confidence levels. The framework’s ability to acquire meaningful augmentation positions it as a novel baseline for semantic-aware confidence calibration.

\section*{Acknowledgements}
This work was supported in part by the Start-up Grant (No. 9610680) of the City University of Hong Kong, Young Scientist Fund (No. 62406265) of NSFC, and the Australian Research Council under Projects DP240101848 and FT230100549.

\section*{Impact Statement}
\begin{comment}
Authors are \textbf{required} to include a statement of the potential 
broader impact of their work, including its ethical aspects and future 
societal consequences. This statement should be in an unnumbered 
section at the end of the paper (co-located with Acknowledgements -- 
the two may appear in either order, but both must be before References), 
and does not count toward the paper page limit. In many cases, where 
the ethical impacts and expected societal implications are those that 
are well established when advancing the field of Machine Learning, 
substantial discussion is not required, and a simple statement such 
as the following will suffice:
\end{comment}
This paper presents work whose goal is to advance the field of 
Machine Learning. There are many potential societal consequences 
of our work, none which we feel must be specifically highlighted here.

\bibliography{example_paper}
\bibliographystyle{icml2025}

%%%%%%%%%%%%%%%%%%%%%%%%%%%%%%%%%%%%%%%%%%%%%%%%%%%%%%%%%%%%%%%%%%%%%%%%%%%%%%%
%%%%%%%%%%%%%%%%%%%%%%%%%%%%%%%%%%%%%%%%%%%%%%%%%%%%%%%%%%%%%%%%%%%%%%%%%%%%%%%
% APPENDIX
%%%%%%%%%%%%%%%%%%%%%%%%%%%%%%%%%%%%%%%%%%%%%%%%%%%%%%%%%%%%%%%%%%%%%%%%%%%%%%%
%%%%%%%%%%%%%%%%%%%%%%%%%%%%%%%%%%%%%%%%%%%%%%%%%%%%%%%%%%%%%%%%%%%%%%%%%%%%%%%
\newpage
\appendix
\onecolumn
{\Large \textbf{Appendix}\vspace{1.0ex}}
\section{Proofs of Equations and Propositions}
\subsection{Proofs of \cref{eq:lambda_expressed}}\label{sec:proofs_reanno}
As we have assumed in advance that a fully optimized model $\operatorname{E}(\cdot)$ should satisfy
\begin{align}
&\lambda=\sigma(\frac{\operatorname{E}(\widetilde{\bm x})^\top\bm P_{i} - \operatorname{E}(\widetilde{\bm x})^\top\bm P_{j}}{\tau}),
\label{eq:probability_ratio_2}
\end{align}
and we can relax the encoded feature and acquire a feature-level mixing coefficient $\lambda_E$ through 
\begin{align}
&\operatorname{E}(\widetilde{\bm x})=\lambda_E\bm P_{i}+(1-\lambda_E)\bm P_{j} + \bm r,\nonumber\\
&s.t.~\bm r^\top(\bm P_{i}-\bm P_{j}) = 0,
\label{eq:feature_expressed_2}
\end{align}
we can first rewrite \cref{eq:feature_expressed_2} by vector multiplying $(\bm P_{i}-\bm P_{j})$ on both sides as
\begin{align}
&\operatorname{E}^\top(\widetilde{\bm x})(\bm P_{i}-\bm P_{j})=\lambda_E\bm P_{i}^\top(\bm P_{i}-\bm P_{j})+(1-\lambda_E)\bm P_{j}^\top(\bm P_{i}-\bm P_{j}) + \bm r^\top(\bm P_{i}-\bm P_{j}),
\end{align}
which can be reformulated as
\begin{align}
(\bm P_{i}-\bm P_{j})^2\lambda_E&=(\operatorname{E}(\widetilde{\bm x})-\bm P_{j})^\top(\bm P_{i}-\bm P_{j})-\bm r^\top(\bm P_{i}-\bm P_{j})\nonumber\\
&=(\operatorname{E}(\widetilde{\bm x})-\bm P_{j})^\top(\bm P_{i}-\bm P_{j})
\end{align}
Therefore, $\lambda_E$ can be expressed as 
\begin{align}
&\lambda_E=\frac{(\operatorname{E}(\widetilde{\bm x})-\bm P_{j})^\top(\bm P_{i}-\bm P_{j})}{(\bm P_{i}-\bm P_{j})^2},
\label{eq:lambda_E_expressed_appendix}
\end{align}
which is independent of $\bm r$. 

Meanwhile, substituting \cref{eq:feature_expressed_2} into \cref{eq:probability_ratio_2}, we can acquire
\begin{align}
\lambda&=\sigma\Big(\frac{1}{\tau}\Big(\lambda_E\bm P_{i}^\top(\bm P_{i}-\bm P_{j})+(1-\lambda_E)\bm P_{j}^\top(\bm P_{i}-\bm P_{j}) + \bm r^\top(\bm P_{i}-\bm P_{j})\Big)\Big)\nonumber\\
&=\sigma\Big(\frac{1}{\tau}\Big((\bm P_{i}-\bm P_{j})^2\lambda_E+\bm P_{j}^\top(\bm P_{i}-\bm P_{j})\Big)\Big)\nonumber\\
&=\sigma\Big(\frac{1}{\tau}\Big((\bm P_{i}-\bm P_{j})^2(\lambda_E-\frac{1}{2})+\frac{1}{2}(\bm P_{i}-\bm P_{j})^2+\bm P_{j}^\top(\bm P_{i}-\bm P_{j})\Big)\Big)\nonumber\\
&=\sigma\Big(\frac{1}{\tau}\Big((\bm P_{i}-\bm P_{j})^2(\lambda_E-\frac{1}{2})+\frac{1}{2}\bm P_{i}^2+\frac{1}{2}\bm P_{j}^2-\bm P_{i}^\top\bm P_{j}+\bm P_{i}^\top\bm P_{j}-\bm P_{j}^2\Big)\Big)\nonumber\\
&=\sigma\Big(\frac{1}{\tau}\Big((\bm P_{i}-\bm P_{j})^2(\lambda_E-\frac{1}{2})+\frac{1}{2}(\bm P_{i}^2-\bm P_{j}^2)\Big)\Big).
\label{eq:lambda_expressed_2}
\end{align}
Therefore, $\lambda_E$ and $\lambda$ can be expressed with \cref{eq:lambda_E_expressed_appendix} and \cref{eq:lambda_expressed_2}, respectively.

\subsection{Proofs of \cref{prop:CE_prop}, \cref{prop:FL_prop}, and \cref{prop:L2_prop}.}\label{sec:proofs_loss}
We start our proofs through restating the definition of a mixup-balanced Loss without the tilde hat for simplicity.

\noindent
\textbf{Mixup-balanced Loss.}
We regard a loss function $\mathcal{L}(\bm p, \bm q): P^{K} \times P^{K} \rightarrow \mathbb R$ as mixup-balanced loss iff the optimal $\bm p^*_1, \bm p^*_2$ for
\begin{align}
&\min_{\bm p_1, \bm p_2}\mathcal{L}(\bm p_1, \bm q_1)+\mathcal{L}(\bm p_2, \bm q_2)\nonumber\\
&~~s.t. ~\|\bm p_1-\bm p_2\|_2^2 \le \delta
\label{eq:problem_appendix}
\end{align}
is a root for score function $\beta(\bm p_1, \bm p_2)=\|\bm p_1-\bm q_1\|_2^2 - \|\bm p_2-\bm q_2\|_2^2$ for all $\delta \ge 0$, where $P^{K}$ is the K-dimensional probability simplex space. $\bm q_i~(i=1,2)$ is the annotated class probability for mixup or perturbed samples of the same source image but with different shifting strengths, \textit{i.e.},
\begin{align}
&\left \{
\begin{array}{l}
\bm x_i=\alpha_i \bm x + (1-\alpha_i) \bm x',\\
\bm q_i=\alpha_i \bm l_s + (1-\alpha_i) \bm l_t,
\end{array}
\right.\nonumber\\
&~~~0.5 \le \alpha_i < 1, i = 1,2,
\label{eq:mixup}
\end{align}
where $\bm l_s$ and $\bm l_t$ are one-hot labels of the source and target classes $s$ and $t$, respectively. The above prediction proximity condition $\|\bm p_1-\bm p_2\|_2^2 < \delta$ is imposed to outline the non-discriminability of correlated mixup samples. Without losing generality, we always assume $0.5 \le \alpha_2 < \alpha_1 <1$ in the following proofs.

\subsection{Proof of \cref{prop:L2_prop}}
The L2 loss is defined as $\mathcal{L}_{L2}(\bm p, \bm q) = \|\bm p-\bm q\|_2^2$, which is mixup-balanced. To prove it, we need to solve
\begin{align}
&\min_{\bm p_1, \bm p_2}\|\bm p_1-\bm q_1\|_2^2+\|\bm p_2-\bm q_2\|_2^2\nonumber\\
&~s.t. ~\|\bm p_1-\bm p_2\|_2^2 \le \delta.
\label{eq:problem_l2}
\end{align}
%Denoting the source and target class indices as $k_1$ and $k_2$, we are solving 

a) When $\delta \ge \|\bm q_1-\bm q_2\|_2^2$, the optimal solution $\bm p_1^*=\bm q_1$ and $\bm p_2^*=\bm q_2$ is mixup-balanced.

b) When $\delta < \|\bm q_1-\bm q_2\|_2^2$, we represent the optima as
\begin{align}
&\left \{
\begin{array}{l}
\bm p_1^*=\lambda_1 \bm q_1 + (1-\lambda_1) \bm q_2 + \bm r_1,\\
\bm p_2^*=\lambda_2 \bm q_2 + (1-\lambda_2) \bm q_1 + \bm r_2,
\end{array}
\right.
\label{eq:l2_reconsider_p_star}
\end{align}
where $\bm r_1^\top(\bm q_1 - \bm q_2) = \bm r_2^\top(\bm q_1 - \bm q_2) = 0$. The objective in Eq. \ref{eq:problem_l2} is reformulated as
\begin{align}
\min_{\bm p_1, \bm p_2}~&\|\bm p_1-\bm q_1\|_2^2+\|\bm p_2-\bm q_2\|_2^2\nonumber\\
=&\| (1-\lambda_1)(\bm q_2-\bm q_1) + \bm r_1 \|_2^2 + \| (1-\lambda_2)(\bm q_1-\bm q_2) + \bm r_2  \|_2^2.
\label{eq:problem_l2_simplified}
\end{align}

Temporarily, we regard $\bm p_i^*$ as vectors in $\mathbb{R}^{K}$ as we'll see the optimal ones always lie in $P^{K}$. With such relaxation, $\bm r_i$ can take any vector value in their subspace, but it optimizes Eq. \eqref{eq:problem_l2_simplified} only when $\bm r_i=\bm 0$, further simplifying Eq. \eqref{eq:problem_l2_simplified} as $\min_{\lambda_1, \lambda_2}~(1-\lambda_1)^2 + (1-\lambda_2)^2$. Substituting Eq. \eqref{eq:l2_reconsider_p_star} into the proximity condition, we get $(\lambda_1+\lambda_2-1)^2 \le \frac{\delta}{\|\bm q_1-\bm q_2\|_2^2}$, \textit{i.e.}, $1 - \sqrt{\frac{\delta}{\|\bm q_1-\bm q_2\|_2^2}} \le \lambda_1+\lambda_2 \le 1 + \sqrt{\frac{\delta}{\|\bm q_1-\bm q_2\|_2^2}}$. For each valid value of $\lambda_1+\lambda_2 = A$, we consider Lagrangian equation
\begin{align}
L(\lambda_1, \lambda_2, \beta)=(1-\lambda_1)^2 + (1-\lambda_2)^2+\beta(\lambda_1+\lambda_2 - A),
\label{eq:lagrangian_l2}
\end{align}
where $\beta$ is the Lagrangian multiplier. By setting the derivative \textit{w.r.t.} $\lambda_i$ to $0$, we solve for the optimum $\lambda_i^*$ as
\begin{align}
\frac{\partial L(\lambda_1, \lambda_2, \beta)}{\partial \lambda_i}\Bigg|_{\lambda_i=\lambda_i^*}&=2(\lambda_i^*-1) + \beta=0\nonumber\\
2(1-\lambda_1^*)&=2(1-\lambda_2^*)=\beta\nonumber\\
\lambda_1^*&=\lambda_2^*=\frac{A}{2} \in (0, 1).
\label{eq:lagrangian_l2_solving}
\end{align}
Therefore, $\bm p_i^*$s are actually interpolations of $\bm q_i$s which definitely lie in $P^K$. We have $\|\bm p_1^* - \bm q_1\|_2^2=\|\bm p_2^* - \bm q_2\|_2^2=(1-\frac{A}{2})\|\bm q_1-\bm q_2\|_2^2$ by substituting Eq. \eqref{eq:lagrangian_l2_solving} and \eqref{eq:l2_reconsider_p_star} into each term, indicating $\forall A, \beta(\bm p_1^*, \bm p_2^*)=0$. Hence, the L2 loss is mixup-balanced.

\subsection{Proofs of \cref{prop:CE_prop} and \cref{prop:FL_prop}}
CE and Focal Losses \textit{are not} Mixup-balanced. We prove that $\beta \le 0$ is always true for the cross entropy loss while for Focal loss with some specific $\gamma$, $\beta \ge 0$ is always the case.

We denote a dual-sample loss function as $\mathcal L^{Pair} = \sum_{i=1}^{2}\sum_{k=1}^{K}q_i^k\mathcal{L}_{Item}(p_i^k)$, where $\mathcal{L}_{Item}(\cdot)$ is monotonous and approaches infinity when $\bm p_i^k$ approaches $0$. To solve the loss minimization objective
\begin{align}
\min_{\bm p_1, \bm p_2}\mathcal L^{Pair}~~s.t. ~\|\bm p_1-\bm p_2\|_2^2 \le \delta, \sum_{k}p_1^k=1, \sum_{k}p_2^k=1,
\label{eq:problem_ce_fl}
\end{align}
we should first clarify several points before looking into it.

a) For every class $k \ne s,t$, $q_i^k=0$, the optimal $p_i^k$ is $0$. Otherwise, we can carry out $3$ types of operations without increasing $\|\bm p_1-\bm p_2\|_2^2$. Op. 1 is formulated as
\begin{align}
p_1^k \leftarrow p_1^k-r, ~~p_2^k \leftarrow p_2^k-r,\nonumber\\
p_1^c \leftarrow p_1^c+r, ~~p_2^c \leftarrow p_2^c+r,
\label{eq:operation_1}
\end{align}
where $c \in \{s, t\}$ and $r > 0$. While this operation is proximity-invariant, it reduces the objective function and can finally decrease at least one of $\{p_i^k, p_j^k\}$ to $0$ for every pair of them indexed by $k$. Then, Op. 2 is given as
\begin{align}
p_1^{k_1} \leftarrow p_1^{k_1}-r, ~~&p_2^{k_2} \leftarrow p_2^{k_2}-r,\nonumber\\
p_1^c \leftarrow p_1^c+r, ~~&p_2^c \leftarrow p_2^c+r,
\label{eq:operation_2}
\end{align}
where $k_1, k_2 \not\in \{s, t\}$, $k_1 \ne k_2$. Op. 2 optimizes both the objective and the proximity value, making only one of $\{\bm p_1, \bm p_2\}$ (denoted as $\bm p_i$) having spare values. Finally, Op. 3 merges these values into $p_i^s$ or $p_i^t$:
\begin{align}
&p_i^k \leftarrow p_i^k-r,\nonumber\\
&p_i^c \leftarrow p_i^c+r,\nonumber\\
\label{eq:operation_3}
\end{align}
where before operation, $p_i^c<p_j^c (i \ne j)$ and $r \le p_j^c-p_i^c$. It's easy to prove the existence of such class $c$ and the reduction of both the objective and the proximity value. 

With these operations, we can optimize the objective when $p_i^k\ne 0$ for any class $k \ne s,t$. Therefore, $p_i^k = 0$ is always true in the optimal $\bm p_i^*$, \textit{i.e.}, ${p^*}_i^s+{p^*}_i^t=1$.

%one can always choose the smallest one in $\{p_1^s, p_1^t, p_2^s, p_2^t\}$, and transfer the value of $p_i^k$ onto the selected variable (within the same sample) to further optimize the objective, while guaranteeing the proximity condition. If there is only one sample having spare values of $p_i^k$, it can still be transferred within the sample $i$ to $p_i^s$ (\textit{resp.} $p_i^t$) which is still less than its counterpart $p_j^s$ (\textit{resp.} $p_j^t$, $i \ne j$). Therefore, we can derive $\sum_{k}p_i^k={p^{*}}_i^s+{p^{*}}_i^t=1$ for $i=1,2$.

b) As we have assumed in advance that $q_2^s=\alpha_2<\alpha_1=q_1^s$, we can now immediately derive that ${p^{*}}_2^s \le {p^{*}}_1^s$. Otherwise, we can swap $\bm {p^{*}}_1$ and $\bm {p^{*}}_2$ to further lower the loss while keeping the proximity constraint unchanged. This helps us to derive
\begin{align}
\|\bm p_1 - \bm p_2\|_2^2&=(p_1^s-p_2^s)^2+(p_1^t-p_2^t)^2\nonumber\\
&=(p_1^s-p_2^s)^2+(1-p_1^s-1+p_2^s)^2\nonumber\\
&=2(p_1^s-p_2^s)^2=A,\nonumber\\
&p_1^{*s}=p_2^{*s}+\sqrt\frac{A}{2}
\label{eq:p1_p2_diff_pre}
\end{align}
for a specific value of $A \in [0, \delta]$. For simplicity, we rewrite Eq. \eqref{eq:p1_p2_diff_pre} as:
\begin{align}
{p^{*}}_1^s={p^{*}}_2^s+\epsilon,
\label{eq:p1_p2_diff}
\end{align}
where $\epsilon \in \Big[0, \sqrt\frac{\delta}{2}\Big]$.

These observations help us formulate and solve the following Lagrangian equation (for arbitrary $A$):
\begin{align}
&L(\bm p_1, \bm p_2, \beta, \beta_1, \beta_2)\nonumber\\
=~&\mathcal L^{Pair}
+\beta(\|\bm p_1 - \bm p_2\|_2^2-A)+\beta_1(\sum_{k}\bm p_1^k-1)
+\beta_2(\sum_{k}\bm p_2^k-1),
\label{eq:lagrangian_ce_fl}
\end{align}
where $\beta$, $\beta_1$ and $\beta_2$ are the Lagrangian multipliers. By taking derivatives \textit{w.r.t.} $p_i^k$ and set it to $0$, we have:
\begin{align}
\frac{\partial L}{\partial p_i^k} = \frac{\partial \mathcal L^{Pair}}{\partial p_i^k} + 2\beta(p_i^k-p_j^k)+\beta_i=0,
\label{eq:lagrangian_ce_fl_solve1}
\end{align}
where $i,j \in \{1,2\}$, $i \ne j$, and $k \in \{s,t\}$. It can be seen that
\begin{align}
&\frac{\partial L}{\partial p_i^s}+\frac{\partial L}{\partial p_i^t}=\frac{\partial \mathcal L^{Pair}}{\partial p_i^s} + \frac{\partial \mathcal L^{Pair}}{\partial p_i^t}+2\beta(p_i^s-p_j^s+p_i^t-p_j^t)+2\beta_i=0\nonumber\\
&\frac{\partial L}{\partial p_i^s}+\frac{\partial L}{\partial p_i^t}=\frac{\partial \mathcal L^{Pair}}{\partial p_i^s} + \frac{\partial \mathcal L^{Pair}}{\partial p_i^t}+2\beta_i=0\nonumber\\
&\beta_i = -\frac{1}{2}\Big(\frac{\partial \mathcal L^{Pair}}{\partial p_i^s} + \frac{\partial \mathcal L^{Pair}}{\partial p_i^t}\Big),
\label{eq:lagrangian_ce_fl_solve_beta_i}
\end{align}
and 
\begin{align}
&\frac{\partial L}{\partial p_1^k}+\frac{\partial L}{\partial p_2^k}=\frac{\partial \mathcal L^{Pair}}{\partial p_1^k} + \frac{\partial \mathcal L^{Pair}}{\partial p_2^k} 
+ 2\beta(p_1^k-p_2^k+p_2^k-p_1^k) +\beta_1+\beta_2=0\nonumber\\
&\frac{\partial L}{\partial p_1^k}+\frac{\partial L}{\partial p_2^k}=\frac{\partial \mathcal L^{Pair}}{\partial p_1^k} + \frac{\partial \mathcal L^{Pair}}{\partial p_2^k} +\beta_1+\beta_2=0.
\label{eq:lagrangian_ce_fl_eq2}
\end{align}
By substituting Eq. \eqref{eq:lagrangian_ce_fl_solve_beta_i} into Eq. \eqref{eq:lagrangian_ce_fl_eq2} and let $k=s$ ($k=t$ yields the same result), we have:
\begin{align}
&\frac{\partial \mathcal L^{Pair}}{\partial p_1^s} + \frac{\partial \mathcal L^{Pair}}{\partial p_2^s} -\frac{1}{2}\Big(\frac{\partial \mathcal L^{Pair}}{\partial p_1^s} + \frac{\partial \mathcal L^{Pair}}{\partial p_1^t}\Big)-\frac{1}{2}\Big(\frac{\partial \mathcal L^{Pair}}{\partial p_2^s} + \frac{\partial \mathcal L^{Pair}}{\partial p_2^t}\Big)=0.\nonumber\\
&~~~~~~~~~~~~~~~~~~~\frac{\partial \mathcal L^{Pair}}{\partial p_1^s}
-\frac{\partial \mathcal L^{Pair}}{\partial p_1^t}
+\frac{\partial \mathcal L^{Pair}}{\partial p_2^s}
-\frac{\partial \mathcal L^{Pair}}{\partial p_2^t}=0.
\label{eq:lagrangian_ce_fl_eq3}
\end{align}

\noindent \textbf{CE Loss.} When $\delta \ge \|\bm q_1-\bm q_2\|_2^2$, the optimal solution in the case of CE loss is $\bm {p^{*}}_1=\bm q_1$ and $\bm {p^{*}}_2=\bm q_2$, which is mixup-balanced. However, when $\delta < \|\bm q_1-\bm q_2\|_2^2$, the CE loss always yield negative $\beta$ scores, indicating its preference for fitting samples close to the source.

We prove this by regarding the left part of Eq. \eqref{eq:lagrangian_ce_fl_eq3} as a function of $p_2^s$ since we have ${p^*}_1^s={p^*}_2^s+\epsilon$ and ${p^{*}}_i^{s}+{p^{*}}_i^{t}=1$. From here on, we discuss the optimal $\bm p_i$ and omit the star superscript for simplicity. The left part of Eq. \eqref{eq:lagrangian_ce_fl_eq3} can be rewritten as
\begin{align}
h(x) = -\frac{\alpha_1}{x+\epsilon}+\frac{1-\alpha_1}{1-x-\epsilon}-\frac{\alpha_2}{x}+\frac{1-\alpha_2}{1-x}.
\label{eq:function_ce}
\end{align}
The root of $h(x)$ in $[0.5, 1-\epsilon]$ is the root $p_2^s$ of Eq. \eqref{eq:lagrangian_ce_fl_eq3}, \textit{i.e.}, ${p^{*}}_2^{s}$. Similarly, the root of $h(x-\epsilon)$ is ${p^{*}}_1^{s}$. Although it is difficult to find a closed form of their roots, we can find the mean of them, \textit{i.e.}, $t=({p^{*}}_1^{s}+{p^{*}}_2^{s})/2$, as the root of $h(x-\epsilon/2)$. Our initial goal is to find the sign of $\beta$, which is now 
\begin{align}
\beta &= (p_1^s-\alpha_1)^2+(1-p_1^s-1+\alpha_1)^2-(p_2^s-\alpha_2)^2-(1-p_2^s-1+\alpha_2)^2\nonumber\\
&= 2(p_1^s-\alpha_1)^2 - 2(p_2^s-\alpha_2)^2\nonumber\\
&= 2(2t-\alpha_1-\alpha_2)(\epsilon+\alpha_2-\alpha_1)
\end{align}
and is closely associated with $t$. The sign of $\beta$ is determined by the sign of $(\alpha_1+\alpha_2)/2-t$. It's intractable to solve $t$ directly, but noting the monotonicity of $f$ with
\begin{align}
h'(x) = \frac{\alpha_1}{(x+\epsilon)^2}+\frac{1-\alpha_1}{(1-x-\epsilon)^2}+\frac{\alpha_2}{x^2}+\frac{1-\alpha_2}{(1-x)^2}>0
\end{align}
for all $x \in (0, 1-\epsilon)$, we now have 
\begin{align}
\frac{\alpha_1+\alpha_2}{2} - t < 0 \Leftrightarrow h(\frac{\alpha_1+\alpha_2}{2}-\frac{\epsilon}{2})<h(t-\frac{\epsilon}{2})=0,
\end{align}
where $h(\frac{\alpha_1+\alpha_2}{2}-\frac{\epsilon}{2})$ equals
\begin{align}
&T_{CE}(\alpha_1, \alpha_2, \epsilon)\nonumber\\
=&-\frac{\alpha_1}{\frac{\alpha_1+\alpha_2}{2}-\frac{\epsilon}{2}+\epsilon}
+\frac{1-\alpha_1}{1-\frac{\alpha_1+\alpha_2}{2}+\frac{\epsilon}{2}-\epsilon}
-\frac{\alpha_2}{\frac{\alpha_1+\alpha_2}{2}-\frac{\epsilon}{2}}
+\frac{1-\alpha_2}{1-\frac{\alpha_1+\alpha_2}{2}+\frac{\epsilon}{2}}\nonumber\\
=&2\Big(\frac{1-\alpha_1}{1-\alpha_1-\alpha_2-\epsilon}-\frac{\alpha_1}{\alpha_1+\alpha_2+\epsilon}+\frac{1-\alpha_2}{1-\alpha_1-\alpha_2+\epsilon}-\frac{\alpha_2}{\alpha_1+\alpha_2-\epsilon}\Big).
\end{align}
Given that $0.5 <= \alpha_2 < \alpha_1 < 1$ and $0 \le\epsilon<\alpha_1-\alpha_2$ (by $\delta < \|\bm q_1-\bm q_2\|_2^2$), we can determine the sign of $T_{CE}(\alpha_1, \alpha_2, \epsilon)$.

\begin{lemma}
\label{lem:CE_inner_lemma}
$T_{CE}(\alpha_1, \alpha_2, \epsilon)<0$. Therefore, when $\delta < \|\bm q_1-\bm q_2\|_2^2$, the CE loss always yields negative $\beta$ scores.
\end{lemma}

\noindent \textit{Proof.} We denote that
\begin{align}
&\left \{
\begin{array}{l}
D_1 = 2 - \alpha_1 - \alpha_2 - \epsilon>0,\\
D_2 = \alpha_1 + \alpha_2 + \epsilon>0,\\
D_3 = 2 - \alpha_1 - \alpha_2 + \epsilon>0,\\
D_4 = \alpha_1 + \alpha_2 - \epsilon>0,
\end{array}
\right.
\end{align}
Then, we have
\begin{align}
\frac{1}{2}T_{CE}(\alpha_1, \alpha_2, \epsilon)=&\frac{1-\alpha_1}{D_1}-\frac{\alpha_1}{D_2}+\frac{1-\alpha_2}{D_3}-\frac{\alpha_2}{D_4}\nonumber\\
=&\frac{(1-\alpha_1)D_2-\alpha_1 D_1}{D_1 D_2} + \frac{(1-\alpha_2)D_4-\alpha_2 D_3}{D_3 D_4}\nonumber\\
=&\frac{-\alpha_1+\alpha_2+\epsilon}{D_1 D_2} + \frac{\alpha_1-\alpha_2-\epsilon}{D_3 D_4}\nonumber\\
=&(\alpha_1-\alpha_2-\epsilon)(\frac{1}{D_3 D_4}-\frac{1}{D_1 D_2})\nonumber\\
=&(\alpha_1-\alpha_2-\epsilon)(\frac{D_1 D_2 - D_3 D_4}{D_1 D_2 D_3 D_4})
\end{align}
where $\alpha_1-\alpha_2-\epsilon> 0$, $D_1 D_2 D_3 D_4>0$, and $D_1 D_2 - D_3 D_4=-4c(a+b-1)<0$. Therefore, $h(\frac{\alpha_1+\alpha_2}{2}-\frac{\epsilon}{2})=T_{CE}(\alpha_1, \alpha_2, \epsilon)<0$. Based on the previous deductions, we can conclude that $\beta$ is always negative when $\delta < \|\bm q_1-\bm q_2\|_2^2$.

\noindent \textbf{Focal Loss.} Due to the complexity of Focal loss, we specifically analyze the case when $\gamma_{FL}=1$. The commonly used $\gamma$s are larger and the balance scores become even greater empirically for these values. It's worth noting that $\delta \ge \|\bm q_1-\bm q_2\|_2^2$ doesn't guarantee the Focal loss to yield $\bm {p^{*}}_i=\bm q_i$ since the Focal loss is not strictly proper \cite{charoenphakdee2021focal}. Nevertheless, its behavior with $\delta < \|\bm q_1-\bm q_2\|_2^2$ can be sufficiently clear. We mainly focus on these cases in the following proofs.

Similar to CE loss, the sign of $\beta$ is associated with function
\begin{align}
g(x) = 
 &(1-\alpha_1)\Big(\frac{x+\epsilon}{1-x-\epsilon}-\log(1-x-\epsilon)\Big)
- \alpha_1    \Big(\frac{1-x-\epsilon}{x+\epsilon}-\log(x+\epsilon)\Big)\nonumber\\
+&(1-\alpha_2)\Big(\frac{x}{1-x}-\log(1-x)\Big)
- \alpha_2    \Big(\frac{1-x}{x}-\log(x)\Big).
\label{eq:function_fl}
\end{align}
Note that $g(x)$ also increases monotonously. Therefore, the sign of $\beta$ at the optimal $\bm p_i^*$ is the same as the sign of $g(\frac{\alpha_1+\alpha_2}{2}-\frac{\epsilon}{2})$, which equals
\begin{align}
T_{FL}(\alpha_1, \alpha_2&,\epsilon) = 
 (1-\alpha_1)\Big(\frac{E}{1-E}-\log(1-E)\Big)
- \alpha_1    \Big(\frac{1-E}{E}-\log(E)\Big)\nonumber\\
+&(1-\alpha_2)\Big(\frac{F}{1-F}-\log(1-F)\Big)
- \alpha_2    \Big(\frac{1-F}{F}-\log(F)\Big),
\label{eq:function_fl_at_mid}
\end{align}
where $E = \frac{\alpha_1+\alpha_2+\epsilon}{2}$, $F=\frac{\alpha_1+\alpha_2-\epsilon}{2}$, constrained by $\alpha_2 < F \le E < \alpha_1$, and $E+F=\alpha_1+\alpha_2$.

\begin{lemma}
\label{lem:FL_inner_lemma}
$T_{FL}(\alpha_1, \alpha_2, \epsilon)>0$. Therefore, when $\delta < \|\bm q_1-\bm q_2\|_2^2$, the FL loss with $\gamma_{FL}=1$ always yields positive $\beta$ scores.
\end{lemma}

\noindent \textit{Proof.} We rewrite $T_{FL}(\alpha_1, \alpha_2,\epsilon)$ as
\begin{align}
&T_{FL}(\alpha_1, \alpha_2,\epsilon) = \mu(\alpha_2,x)+\mu(\alpha_1,-x),\nonumber\\
&\mu(w,x)=(1-w)\Big(\frac{w+x}{1-w-x}-\log(1-w-x)\Big)\nonumber\\
&~~~~~~~~~~~~~~~-w\Big(\frac{1-w-x}{w+x}-\log(w+x)\Big),
\label{eq:function_fl_at_mid_rewrite}
\end{align}
where $w \in [0.5, 1.0)$ and $x \in (0, \min(\frac{\alpha_1-\alpha_2}{2}, 1-w)]$. We show 2 crucial properties of $\mu(w,x)$:

a) $\mu(w,-x) + \mu(0.5,x) > 0$. 

\textit{Proof.} We reformulate the left part of the inequality as
\begin{align}
&\mu(w,-x) + \mu(0.5,x)\nonumber\\
=&(1-w)\Big(\frac{w-x}{1-w+x}-\log(1-w+x)\Big)-w\Big(\frac{1-w+x}{w-x}-\log(w-x)\Big)\nonumber\\
&+0.5\Big(\frac{0.5+x}{0.5-x}-\log(0.5-x)\Big)-0.5\Big(\frac{0.5-x}{0.5+x}-\log(0.5+x)\Big)\nonumber\\
=&-(1-w)\log(1-w+x)+w\log(w-x)-0.5\log(0.5-x)+0.5\log(0.5+x)\nonumber\\
&+(1-w)\frac{w-x}{1-w+x}-w\frac{1-w+x}{w-x}+0.5\frac{0.5+x}{0.5-x}-0.5\frac{0.5-x}{0.5+x},
\label{eq:property_1_expanded}
\end{align}
where we consider
\begin{align}
&T_1=w\log(w-x)-0.5\log(0.5-x),\\
&T_2=0.5\log(0.5+x)-(1-w)\log(1-w+x).
\end{align}
to find that
\begin{align}
&\frac{\partial T_1}{\partial x}=\frac{0.5}{0.5-x}-\frac{w}{w-x}=\frac{(w-0.5)x}{(0.5-x)(w-x)} \ge 0,\\
&\frac{\partial T_2}{\partial x}=\frac{0.5}{0.5+x}-\frac{1-w}{1-w+x}=\frac{(w-0.5)x}{(0.5+x)(1-w+x)} \ge 0.\\
&T_1+T_2 > (T_1+T_2)\big|_{x=0}=\phi(w)=w\log w-(1-w)\log(1-w).
\end{align}
Because $\phi''(w)=\frac{1-2w}{w(1-w)}\le 0$, $T_1+T_2 > \phi(w)\ge \min(\phi(0.5), \phi(1^-))=0$.
Now that we have considered the sum of logarithmic terms in Eq. \eqref{eq:property_1_expanded}, we proceed to inspect the sum of the rest terms:
\begin{align}
&(1-w)\frac{w-x}{1-w+x}-w\frac{1-w+x}{w-x}+0.5\frac{0.5+x}{0.5-x}-0.5\frac{0.5-x}{0.5+x}\nonumber\\
>&(1-w)\frac{0.5-x}{0.5+x}-w\frac{1-w+x}{w-x}+0.5\frac{0.5+x}{0.5-x}-0.5\frac{0.5-x}{0.5+x}\nonumber\\
=&F(w,x)
\end{align}
where we can find
\begin{align}
\frac{\partial F(w,x)}{\partial w}&=-\frac{0.5-x}{0.5+x}-\frac{1-w+x}{w-x}-w\frac{x-w-(1-w+x)}{(w-x)^2}\nonumber\\
&=-\frac{0.5-x}{0.5+x}-\frac{1-w+x}{w-x}+\frac{w}{(w-x)^2}\nonumber\\
&=\frac{w-(w-x)(1-w+x)}{(w-x)^2}-\frac{0.5-x}{0.5+x}\nonumber\\
&=\frac{x+(w-x)^2}{(w-x)^2}-\frac{0.5-x}{0.5+x}\nonumber\\
&=1+\frac{x}{(w-x)^2}-\Big(1-\frac{2x}{0.5+x}\Big)\nonumber\\
&=\frac{x}{(w-x)^2}+\frac{2x}{0.5+x}>0\\
F(w,x) &> F(0.5, x)=0.
\end{align}
Therefore, Eq. \eqref{eq:property_1_expanded} is positive because the sum of all its terms is positive.

b) $\mu(w,x) > \mu(0.5,x)$

\textit{Proof.} 
\begin{align}
&\mu(w,x) - \mu(0.5,x)\nonumber\\
=&-(1-w)\log(1-w-x)+w\log(w+x)+0.5\log(0.5-x)-0.5\log(0.5+x)\nonumber\\
&+(1-w)\frac{w+x}{1-w-x}-w\frac{1-w-x}{w+x}-0.5\frac{0.5+x}{0.5-x}+0.5\frac{0.5-x}{0.5+x},
\label{eq:property_2_expanded}
\end{align}
where we consider
\begin{align}
T_3=&-(1-w)\log(1-w-x)+w\log(w+x)\nonumber\\
&+0.5\log(0.5-x)-0.5\log(0.5+x),\\
T_4=&(1-w)\frac{w+x}{1-w-x}-w\frac{1-w-x}{w+x}-0.5\frac{0.5+x}{0.5-x}+0.5\frac{0.5-x}{0.5+x}.
\end{align}
We can find that
\begin{align}
\frac{\partial T_3}{\partial x}=&\frac{1-w}{1-w-x}+\frac{w}{w+x}-\frac{0.5}{0.5-x}-\frac{0.5}{0.5+x}\nonumber\\
=&\frac{0.5(1-w)-x(1-w)-0.5(1-w)+0.5x}{(1-w-x)(w+x)}+\frac{0.5w+wx-0.5w-0.5x}{(w+x)(0.5+x)}\nonumber\\
=&\frac{(w-0.5)x}{(1-w-x)(w+x)}+\frac{(w-0.5)x}{(w+x)(0.5+x)}\nonumber\\
>&0\\
T_3>&T_3\big|_{x=0}=w\log w-(1-w)\log(1-w)>0.
\end{align}
Meanwhile, we can also find
\begin{align}
T_4=&(1-w)\frac{w+x}{1-w-x}-w\frac{1-w-x}{w+x}-0.5\frac{0.5+x}{0.5-x}+0.5\frac{0.5-x}{0.5+x}\nonumber\\
>&(1-w)\frac{w+x}{1-w-x}-w\frac{0.5-x}{0.5+x}-0.5\frac{0.5+x}{0.5-x}+0.5\frac{0.5-x}{0.5+x}\nonumber\\
=&G(w,x).
\end{align}
where we have
\begin{align}
\frac{\partial G(w,x)}{\partial w}&=-\frac{w+x}{1-w-x}+(1-w)\frac{1-w-x+w+x}{(1-w-x)^2}-\frac{0.5-x}{0.5+x}\nonumber\\
&=\frac{1-w-(w+x)(1-w-x)}{(1-w-x)^2}-\frac{0.5-x}{0.5+x}\nonumber\\
&=\frac{(1-w-x)^2+x}{(1-w-x)^2}-\frac{0.5-x}{0.5+x}\nonumber\\
&=\frac{x}{(1-w-x)^2}+\frac{2x}{0.5+x}>0,
\end{align}
which results in $T_4>G(w,x)> G(0.5, x)=0$.

Therefore, combining a) and b), we have $-\mu(\alpha_1,-x) < \mu(0.5,x) < \mu(\alpha_2,x)$, \textit{i.e.}, $T_{FL}(\alpha_1, \alpha_2,\epsilon)=\mu(\alpha_2,x)+\mu(\alpha_1,-x)> 0$.

\section{More Details for Experiments}\label{sec:more_exp_info}
\subsection{Compared Baselines} 
We adopt diverse training-time methods for comparison including the vanilla CE loss. We set hyperparameters for the compared methods following \cite{noh2023rankmixup}. Specifically, we compare calibration performance with the traditional regularization-based methods including 
a) ECP \cite{pereyra2017regularizing} with 0.1 as the coefficient for entropy penalty, 
b) MMCE \cite{kumar2018trainable}, 
c) LS \cite{muller2019does} with $\alpha = 0.05$, 
d) mbLS \cite{liu2022devil} with $m=6$ for CIFAR10/100 and $m=10$ for Tiny-ImageNet, 
e) FL \cite{focal_loss} with fixed $\gamma=3$, 
f) FLSD \cite{FLSD} adopting $\gamma$ schedule of FLSD-53 variant, 
g) CPC \cite{cheng2022calibrating}, 
and h) FCL \cite{liang2024calibrating} with $\gamma=3$ and $\lambda=0.5$. 
We also compare our CSM with data-driven calibration methods, \ie, 
a) Mixup \cite{zhang2018mixup} with a Beta distribution shape parameter $\alpha=0.2$, 
b) RegMixup \cite{pinto2022using} with a Beta distribution shape parameter $\alpha=10$, 
c) AugMix \cite{hendrycks2019augmix} with a coefficient of $1.0$ for JS Divergence term, which yields better accuracy compared to original value of $12$,
and d) RankMixup (M-NDCG) \cite{noh2023rankmixup} with weight for M-NDCG set as $0.1$ and shape parameters $\alpha=1,2$ for CIFAR10/100 and TinyImageNet, respectively. 

\subsection{Datasets and Augmented Samples} 
We adopt CIFAR-10, CIFAR-100, and Tiny-ImageNet as the evaluated datasets. 
We hold the principle that the ratios of generated sample amount versus training size are fixed to be less than a certain value across different datasets, \eg, $4.0$ in our experiments. 
Note that the number of augmented samples we use in training is much less than that in Mixup methods. 
Meanwhile, we fix the size of each sample set generated with the same noise $\bm x^{T}$ as 8 for all datasets. 
Here are detailed descriptions of the datasets and augmented samples:

\textbf{CIFAR-10}: The CIFAR-10 dataset contains $60,000$ RGB images of the size $32 \times 32$. All images fall into one of 10 semantic categories. By default, the dataset is split as $50,000$, $5,000$, and $5,000$ samples for training, validation, and testing. We generate $550$ sets for each of the $45$ class pairs, producing $198,000$ samples for confidence-aware augmentation.

\textbf{CIFAR-100}: The CIFAR-100 dataset consists of $60,000$ $32 \times 32$ color images in $100$ classes. The split is similar to CIFAR-10 with $50,000$ for training, $5,000$ for validation, and $5,000$ for testing. We generate $5$ sample sets for each distinctive class pair, yielding $198,400$ augmented samples in total. 

\textbf{Tiny-ImageNet}: The Tiny-ImageNet dataset is a subset of the large-scale ImageNet dataset and includes $120,000$ images of a large set of $200$ classes. Each image is a downsized ImageNet sample of size $64 \times 64$. For CSM augmentation, we generate $2$ sets for every $19,900$ class pair, producing a total of $318,400$ samples. 

The generation process takes as much or even less time than the training time. Specifically, it takes less than 4h for a single A4000 GPU to generate the amount of all CIFAR-10 or CIFAR-100 augmented samples, while using less than 8h for the same computing units to generate for Tiny-ImageNet. 

\subsection{Evaluation Metrics} 
To assess the model calibration, we employ four key metrics: Expected Calibration Error (ECE), Adaptive Expected Calibration Error (AECE), Overconfidence Error (OE), and Underconfidence Error (UE).

ECE approximates the average discrepancy between a model’s confidence and accuracy. Practically, it is estimated across equally spaced confidence bins. Let $B_1, \dots, B_M$ denote $M$ bins partitioning predictions into intervals $[0, \frac{1}{M}), \dots, [\frac{M-1}{M}, 1]$. For each bin $B_m$, the accuracy and confidence are aomputed as $\text{acc}(B_m) = \frac{1}{|B_m|} \sum_{i \in B_m} \mathbb{I}(y_i = \hat{y}_i)$ and $\text{conf}(B_m) = \frac{1}{|B_m|} \sum_{i \in B_m} \hat{p_i}$, respectively, where $y_i$ is the true label, $\hat{y}_i$ is the predicted label, and $\hat{p_i}$ is the predicted probability for the winning class. ECE is then defined as
\begin{align}
\text{ECE} = \sum_{m=1}^M \frac{|B_m|}{N} \left| \text{acc}(B_m) - \text{conf}(B_m) \right|,
\label{ECE_appendix}
\end{align}
where $N$ is the total number of samples. Lower ECE values indicate better calibration.

AECE addresses potential biases from fixed-width binning by constructing bins with adaptive widths to ensure equal \textit{sample counts} per bin. This approach reduces sensitivity to irregular confidence distributions. The calculation mirrors ECE by using Eq. \eqref{ECE_appendix} but uses bins $B_1, \dots, B_M$ where each bin contains approximately $N/M$ samples. 

OE isolates cases where the model’s confidence exceeds its accuracy compared to ECE. Using the same binning as ECE, OE is defined as
\begin{align}
\text{OE} = \sum_{m=1}^M \frac{|B_m|}{N} \max\left(0, \text{conf}(B_m) - \text{acc}(B_m)\right).
\end{align}
This metric evaluates overconfident predictions, with lower values indicating fewer instances of excessive confidence.

Conversely, UE captures scenarios where the model’s confidence underestimates its accuracy. It is computed as
\begin{align}
\text{UE} = \sum_{m=1}^M \frac{|B_m|}{N} \max\left(0, \text{acc}(B_m) - \text{conf}(B_m)\right).
\end{align}
Lower UE values suggest better alignment in underconfident cases.

\section{More Experimental Results}\label{sec:more_exp_results}
\textbf{Network Architecture~}
We provide results on Wide-ResNet-26-10 and DenseNet-121 to verify the performance consistency of our method in \cref{table_wide_resnet} and \cref{table_densenet}.

\begin{table}[!t]
\begin{center}
\caption{Evaluation results with Wide-ResNet-26-10 on CIFAR-10 and Tiny-ImageNet.} \label{table_wide_resnet}
\resizebox{0.48\textwidth}{!}{
%\begin{tabular}{c|c|ccc|ccc|ccc|ccc}
%\begin{tabular}{llcccccccccccc}
\begin{tabular}{lcccccc}
\toprule
\multicolumn{1}{c}{\multirow{3}{*}{Method}} & \multicolumn{3}{c}{CIFAR-10}                                            & \multicolumn{3}{c}{Tiny-ImageNet}                                      \\
\multicolumn{1}{c}{}                        & \multicolumn{6}{c}{Wide-ResNet-26-10}                                                                                                            \\
\multicolumn{1}{c}{}                        & ACC$\uparrow$  & ECE$\downarrow$ & \multicolumn{1}{l}{AECE$\downarrow$} & ACC$\uparrow$ & ECE$\downarrow$ & \multicolumn{1}{l}{AECE$\downarrow$} \\ 
\midrule
CE                                          & 95.80          & 2.70            & 2.66                                 & 65.18         & 6.08            & 6.06                                 \\
Mixup                                       & \textbf{96.53} & 3.14            & 3.08                                 & 66.36         & 3.77            & 3.75                                 \\
MbLS                                        & 95.70          & 1.45            & 2.78                                 & 65.30         & 2.57            & 2.32                                 \\
RegMixup                                    & 95.44          & 4.18            & 3.99                                 & 63.40         & 3.87            & 3.93                                 \\
FCL  & 95.84 & 0.92 & 1.39 & 64.62	 & 6.85 & 6.85 \\
RankMixup                                   & 95.73          & 1.62            & 1.53                                 & 65.56         & 3.83            & 3.94                                 \\
\rowcolor{gray!12}
CSM                                        & 96.09          & \textbf{0.49}   & \textbf{0.23}                        & \textbf{67.81}             & \textbf{1.67}               & \textbf{1.66}                                    \\ 
\bottomrule
\end{tabular}
}
\end{center}
\end{table}

\begin{table}[!t]
\begin{center}
\caption{Evaluation results with DenseNet-121 on CIFAR-10.} \label{table_densenet}
\resizebox{0.48\textwidth}{!}{
%\begin{tabular}{c|c|ccc|ccc|ccc|ccc}
%\begin{tabular}{llcccccccccccc}
\begin{tabular}{lccccccc}
\toprule
Metric           & \multicolumn{1}{c}{CE} & \multicolumn{1}{c}{MbLS} & \multicolumn{1}{c}{Mixup} & \multicolumn{1}{c}{RegMixup} & \multicolumn{1}{c}{FCL} & \multicolumn{1}{c}{RankMixup} & \multicolumn{1}{c}{CSM} \\ 
\midrule
ACC$\uparrow$    & 94.68                  & 94.91                    & 95.28                           & \textbf{96.23}                 & 95.41                   & 94.66                         & 95.71                   \\
ECE$\downarrow$  & 3.37                   & 1.64                     & 2.01                           & 5.60                         & 0.66                    & 2.87                          & \textbf{0.51}                    \\
AECE$\downarrow$ & 3.31                   & 3.52                     & 2.15                           & 5.39          & 1.28                    & 2.84                          & \textbf{0.19}                    \\ 
\bottomrule
\end{tabular}
}
\end{center}
\end{table}

\begin{table}[!t]
\begin{center}
\caption{Results on long-tailed datasets.} \label{table_long_tailed}
\resizebox{0.48\textwidth}{!}{
%\begin{tabular}{c|c|ccc|ccc|ccc|ccc}
%\begin{tabular}{llcccccccccccc}
\begin{tabular}{lcccccccc}
\toprule
\multicolumn{1}{c}{\multirow{3}{*}{Method}} & \multicolumn{4}{c}{CIFAR10-LT}                                      & \multicolumn{4}{c}{CIFAR100-LT}                                     \\
\multicolumn{1}{c}{}                         & \multicolumn{2}{c}{$\rho=10$}    & \multicolumn{2}{c}{$\rho=100$}   & \multicolumn{2}{c}{$\rho=10$}    & \multicolumn{2}{c}{$\rho=100$}   \\
\multicolumn{1}{c}{}                         & ACC$\uparrow$  & ECE$\downarrow$ & ACC$\uparrow$  & ECE$\downarrow$ & ACC$\uparrow$  & ECE$\downarrow$ & ACC$\uparrow$  & ECE$\downarrow$ \\ 
\midrule
CE                                           & 86.39          & 6.60            & 70.36          & 20.53           & 55.70          & 22.85           & 38.32          & 38.23           \\
Mixup                                        & 87.10          & 6.55            & 73.06          & 19.20           & 58.02          & 19.69           & 39.54          & 32.72           \\
Remix                                        & 88.15          & 6.81            & 75.36          & 15.38           & 59.36          & 20.17           & 41.94          & 33.56           \\
UniMix                                       & 89.66          & 6.00            & 82.75          & 12.87           & 61.25          & 19.38           & 45.45          & 27.12           \\
RankMixup                                    & 89.80          & 5.94            & 75.41          & 14.10           & \textbf{63.83}          & 9.99            & 43.00          & 18.74           \\
\rowcolor{gray!12}
CSM                                         & \textbf{90.97} & \textbf{1.81}   & \textbf{86.22} & \textbf{3.73}   & 62.35 & \textbf{7.45}   & \textbf{48.91} & \textbf{16.02}   \\ 
\bottomrule
\end{tabular}
}
\end{center}
\end{table}

\textbf{Long-tailed Datasets~}
We also analyze the effectiveness of CSM on long-tailed datasets, where the challenges of class imbalance and miscalibration are more severe. Following the commonly-adopted setup in \cite{xu2021towards, zhong2021improving}, ResNet32 network and CIFAR10/100-LT datasets are adopted as the backbone and benchmarks, respectively. 
As displayed in \cref{table_long_tailed}, our method outperforms the vanilla mixup and existing mixup-based LT approaches \cite{chou2020remix, xu2021towards, noh2023rankmixup} across datasets and imbalance factors ($\rho=10$ and $\rho=100$) in terms of both ACC and ECE, demonstrating competitive or superior effectiveness without particular designs to learn on long-tailed datasets. 
These results suggest that our model's authentic augmentations and confidence-aware data are key factors contributing to its superior performance, even in highly imbalanced LT settings. Our model's significant improvements in both accuracy and calibration metrics demonstrate its robustness in handling long-tailed datasets.

\begin{figure*}[ht]
\vskip 0.2in
\begin{center}
\centerline{
\subfigure[]{
    \includegraphics[height=0.26\textwidth]{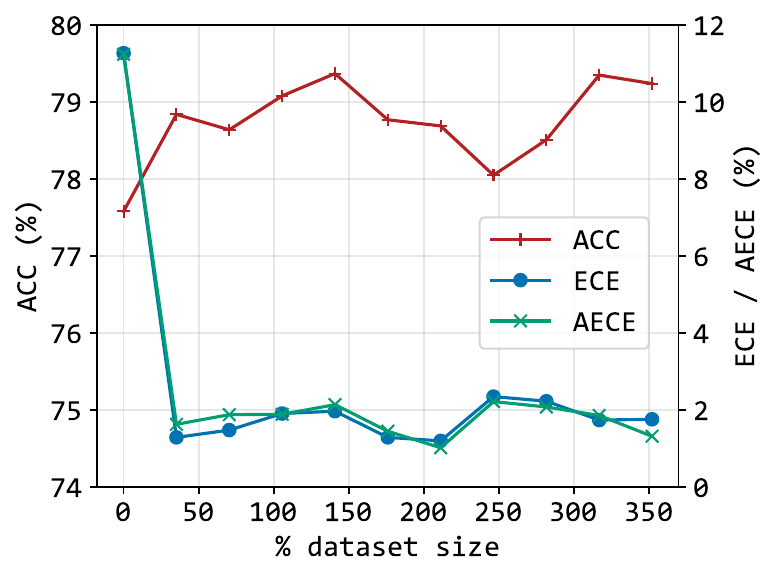}
}
}
\caption{\textbf{Analysis of the amount of augmented data}: Classification ACC, ECE, and AECE on CIFAR-100 \textit{w.r.t} different number of augmented samples relative to the dataset size.}
\label{parameters_nsample}
\end{center}
\vskip -0.2in
\end{figure*}

\begin{table}[!t]
\begin{center}
\caption{\textbf{Calibration Comparison with Different Number of Augmented Samples \textit{per Training Sample.}}} \label{table_aug_per_train}
\vskip 0.1in
\begin{center}
\begin{small}
\begin{sc}
\resizebox{0.36\textwidth}{!}{
\begin{tabular}{lccc}
\toprule
$N_{\text{augs}}$           & 1 & 2 & 3 \\ 
\midrule
ECE: CIFAR-10  &  0.83  &  0.54  &  0.39  \\
ECE: CIFAR-100  &  2.07  &  1.29  &  1.74 \\
\bottomrule
\end{tabular}
}
\end{sc}
\end{small}
\end{center}
\vskip -0.1in
\end{center}
\end{table}

\begin{table}[!t]
\begin{center}
\caption{\textbf{Comparison of the Estimated Training Time.}} \label{table_est_tt}
\vskip 0.1in
\begin{center}
\begin{small}
\begin{sc}
\resizebox{0.80\textwidth}{!}{
\begin{tabular}{lccccccc}
\toprule
Method           & CE  &  MbLS & Mixup & RegMixup & RankMixup & Ours & Ours(EQ-DATA)        \\ 
\midrule
Training Time  &  2.63h & 2.65h & 3.48h & 3.50h & 4.30h & 4.28h & 2.64h \\
\bottomrule
\end{tabular}
}
\end{sc}
\end{small}
\end{center}
\vskip -0.1in
\end{center}
\end{table}

\textbf{Number of Augmented Samples~}
We plot the calibration results \textit{w.r.t.} different amount of used augmentations in \cref{parameters_nsample}(c). To ensure equal comparison, we randomly sample dataset samples and augmented ones with the fixed ratio 1:2 during training regardless of the available amount of augmentations. Surprisingly, the model can calibrate well even with relatively small proportions of augmentations. Meanwhile, ECE and AECE values are jointly minimized with augmentation amount around $200\%$ dataset size, reaching their best value of $1.20$ and $1.02$ on CIFAR-100. These results effectively validate the consistency and reliability of our method.

\textbf{Number of Augmented Samples per Training Sample}
We provide more analyses about the number of augmented samples \textit{per training sample} (denoted as $N_{aug}$) with results in \cref{table_aug_per_train}.
It can be observed that adding the number of accompanied augmentations per dataset sample can generally improve the final calibration performance. This is because a larger number of $N_{augs}$ can sample more sufficient proximal data for training, better filling the domain space and providing more accurate confidence estimation. However, simply using larger $N_{augs}$s could also raise the computational overhead and slow down the training.

\textbf{Estimated Training Time}
We provide a comparison of the estimated training time in \cref{table_est_tt}. One can see the number of augmented samples per batch is the main factor influencing the training time. CSM outperforms others in calibration while maintaining reasonable speed. Even with equalized training samples, \ie, \textsc{EQ-DATA}, it achieves competitive calibration performance. Augmented samples need no re-generation across model, objective, or annotation changes, enabling efficient modular study. We run CSM with a single RTX A4000 device.

\begin{comment}\section{You \emph{can} have an appendix here.}

You can have as much text here as you want. The main body must be at most $8$ pages long.
For the final version, one more page can be added.
If you want, you can use an appendix like this one.  

The $\mathtt{\backslash onecolumn}$ command above can be kept in place if you prefer a one-column appendix, or can be removed if you prefer a two-column appendix.  Apart from this possible change, the style (font size, spacing, margins, page numbering, etc.) should be kept the same as the main body.
\end{comment}
%%%%%%%%%%%%%%%%%%%%%%%%%%%%%%%%%%%%%%%%%%%%%%%%%%%%%%%%%%%%%%%%%%%%%%%%%%%%%%%
%%%%%%%%%%%%%%%%%%%%%%%%%%%%%%%%%%%%%%%%%%%%%%%%%%%%%%%%%%%%%%%%%%%%%%%%%%%%%%%

\end{document}